\documentclass{article}

\usepackage[preprint]{main}

\usepackage[utf8]{inputenc} 
\usepackage[T1]{fontenc}    
\usepackage{hyperref}       
\usepackage{url}            
\usepackage{booktabs}       
\usepackage{amsfonts}       
\usepackage{nicefrac}       
\usepackage{microtype}      
\usepackage[dvipsnames, table]{xcolor}         
\usepackage{amsmath}
\usepackage{amssymb}
\usepackage{mathtools}
\usepackage{amsthm}
\usepackage{cleveref}
\usepackage{enumitem}

\theoremstyle{plain}
\newtheorem{theorem}{Theorem}

\theoremstyle{definition}

\newtheorem{assumption}{Assumption}
\theoremstyle{remark}

\newtheorem{prooft}{Proof}
\usepackage{algorithm}
\usepackage{algorithmic}

\usepackage{multirow}

\usepackage{makecell}
\usepackage{pifont}

\def\algname{{\tt EMA-Nesterov}}
\def\bx{\mathbf{x}}
\def\by{\mathbf{y}}
\def\bm{\mathbf{m}}
\def\bv{\mathbf{v}}
\def\btheta{\mathbf{\theta}}
\newcommand{\dotp}[2]{\left\langle{#1}\ \middle|\ {#2}\right\rangle}

\title{EMA-Nesterov: Stabilizing Nesterov's Lookahead for Accelerated Deep Learning Optimization}

\author{%
  Chung-Yiu Yau$^1$ \And
  Dawei Li$^{1*}$ \And
  Athanasios Glentis$^{1}$\thanks{Equally contributing second author.}
  \AND
  Valentyn Boreiko$^2$\thanks{This work is unrelated to Valentyn Boreiko's position at Amazon.} \And
  Hoi-To Wai$^3$ \And
  Mingyi Hong$^1$ \\ \\
  $^1$University of Minnesota \qquad
  $^2$Amazon AGI \qquad
  $^3$The Chinese University of Hong Kong \\
  $^1$\texttt{\{cyau,li004678,glent007,mhong\}@umn.edu} \\
  $^2$\texttt{valentyn.boreiko@gmail.com} \qquad
  $^3$\texttt{htwai@se.cuhk.edu.hk}
}

\begin{document}

\maketitle

\begin{abstract}

Lookahead-based acceleration methods, such as Nesterov’s momentum, are widely used in optimization, but they often become unreliable in deep learning training mainly due to stochastic gradient noise and non-convex loss landscapes. In particular, standard lookahead relies on short-horizon update signals (e.g., differences between consecutive iterates), which are inherently noisy and can lead to unstable extrapolation directions.
This work revisits Nesterov’s acceleration from a trajectory perspective and argues that effective acceleration in deep learning should harness the low-frequency trends of optimization trajectories rather than extrapolating noisy one-step updates. Leveraging this insight, we propose {\tt EMA-Nesterov}, a simple modification that replaces the standard Nesterov’s lookahead direction with an exponential moving average (EMA) of parameter updates. This yields a stabilized lookahead direction that captures and harnesses the evolving trend of the training trajectory through a low-pass filter, while remaining adaptive to progressive changes via the geometric weighting structure of EMA.
We show that {\tt EMA-Nesterov} retains a theoretical accelerated convergence rate in convex problems that is analogous to Nesterov's accelerated gradient method. Furthermore, we provide empirical evidence on language model pre-training to verify that {\tt EMA-Nesterov} is broadly applicable across a range of fine-tuned base optimizers, including Adam, SOAP, Muon, as well as complex optimizers that achieve state-of-the-art performance on optimization benchmarks (NanoGPT). Compared to prior lookahead methods, {\tt EMA-Nesterov} achieves better performance by avoiding the instability of short-horizon lookahead and the non-adaptivity of long-horizon lookahead.
\end{abstract}
\vspace{-0.5\baselineskip}
\section{Introduction}
\vspace{-0.5\baselineskip}
The growing scale of deep learning training has motivated a new generation of optimizers, such as Adam \citep{kingma2014adam}, SOAP \citep{vyas2024soap}, and Muon \citep{jordan2024muon}, which improve optimization through adaptive preconditioning and  normalization mechanisms that regularize stochastic gradient updates and stabilize training in highly irregular non-convex loss landscapes. 
Beyond the (base) optimizer design, there has also been renewed interest in incorporating trajectory-level momentum into deep learning optimization algorithms. In particular, recent works revisited Nesterov's accelerated gradient method \citep{nesterov1983method}, which extrapolates optimization trajectories using past iterate information (a.k.a.\ lookahead) to achieve accelerated convergence in convex optimization. Moreover, acceleration phenomena have also been theoretically established beyond convex settings: \citet{hinder2020near,gupta2024nesterov} showed acceleration for certain non-convex problems, while \citet{vaswani2019fast} proved accelerated convergence rates for SGD under over-parameterized regimes. 

\begin{figure}[t]
    \centering
    \includegraphics[width=0.9\linewidth]{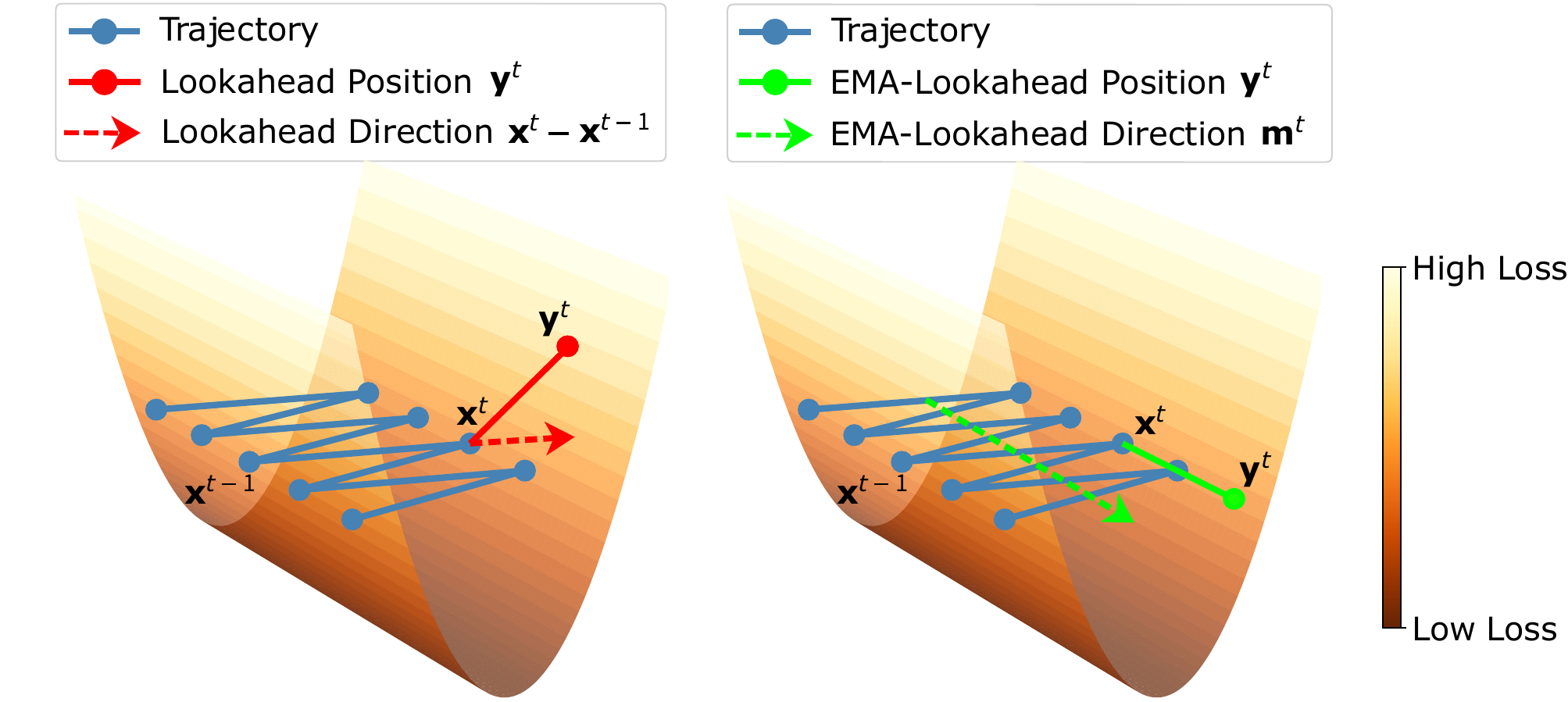}
    \caption{A hypothetical illustration of the training trajectory $(x_1, x_2, f(x_1, x_2)) \in \mathbb{R}^3$, assuming river valley loss landscape and a training trajectory under the WSD learning rate scheduling \citep{wen2024understanding}. We compare the function value of the lookahead position at (Left) $\by^t = \bx^t + \beta \cdot (\bx^{t} - \bx^{t-1})$ which arrives at a high loss position (i.e., $f(y_1^t, y_2^t) > f(x_1^t, x_2^t)$) and (Right) $\by^t = \bx^t + \beta \cdot {\bf m}^t$ which progresses towards the trend of training trajectory where ${\bf m}^t = \gamma {\bf m}^{t-1} + (1-\gamma) (\bx^t - \bx^{t-1})$ for $\gamma \in (0,1)$. }
    \label{fig:river_valley}
\end{figure}
    
Generally speaking, lookahead acceleration constructs an extrapolated solution by moving beyond the current iterate along a direction induced by past optimization trajectories. In the classical Nesterov's accelerated gradient method, this extrapolation direction is given by the difference between consecutive iterates, i.e., $\mathbf{x}^t - \mathbf{x}^{t-1}$. 
One key insight behind why such lookahead direction can accelerate base optimizers is that extrapolating past iterates allows the algorithm to follow the persistent trend of the optimization trajectory, thereby moving more aggressively along stable descent directions. However, under the deep learning optimization trajectory, Nesterov's lookahead can be unstable due to oscillating optimizer update that overshoots the minima under pre-conditioning, as similarly observed in stochastic optimization \citep{wang2022scheduled}. In fact, such a lookahead has not been widely used in contemporary large-scale deep neural network training. 
Instead, exponential moving average of the past (stochastic) gradients has been a key component for modern optimizers such as Adam and Muon.
Muon uses the name Nesterov in its implementation\footnote{According to Muon's open-sourced implementation at \url{https://github.com/KellerJordan/Muon}.} but our careful inspection in Appdenix \ref{app:muon} shows that it is actually an EMA of stochastic gradient with varying EMA rate, and has no connection to Nesterov's momentum or Nesterov's lookahead.

\begin{figure}[htb!]
    \centering
    \includegraphics[width=0.4\linewidth]{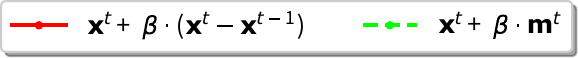} \\
    \includegraphics[width=0.325\linewidth]{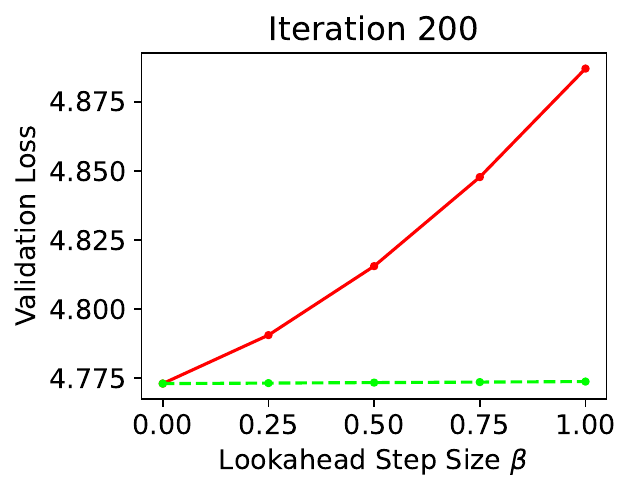}
    \includegraphics[width=0.325\linewidth]{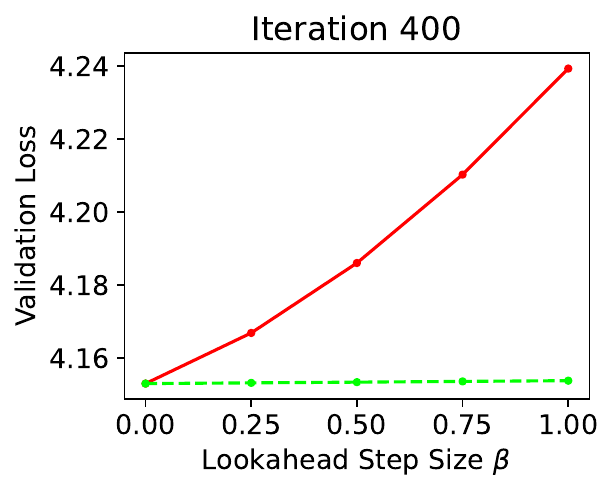}
    \includegraphics[width=0.325\linewidth]{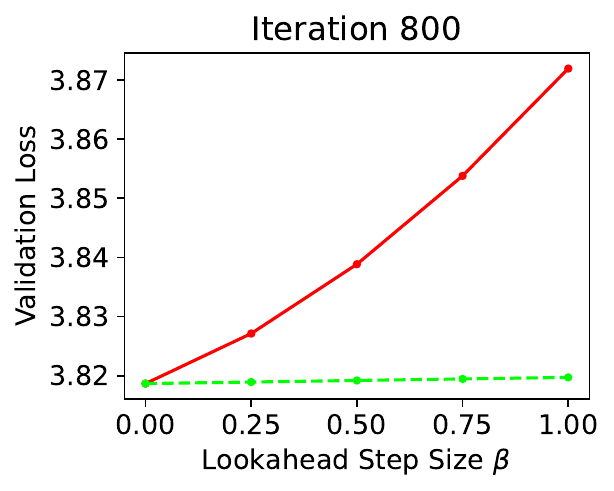}
    \caption{Under the NanoGPT setup, we measure the validation loss in different lookahead positions ${\bf y}^t$ at iterations $t=200, 400, 800$ using ({\color{red}{Solid Red}}) $\by^t = \bx^t + \beta \cdot (\bx^t - \bx^{t-1})$ and ({\color{green!50!black}Dashed Green}) $\by^t = \bx^t + \beta \cdot {\bf m}^t$ where ${\bf m}^t = \gamma {\bf m}^{t-1} + (1-\gamma) (\bx^t - \bx^{t-1})$ with $\gamma=0.995$. The usual lookahead leads to sharp increase in loss and EMA finds stabilized lookahead direction to avoid a sharp increase. 
    }
    \label{fig:nanogpt_lookahead}
\end{figure}

To gain intuition on the training trajectory of deep learning optimization and how lookahead directions affect such trajectories, in Figure \ref{fig:river_valley} we consider a simplified river-valley landscape, in which during the stable phase of training, the iterate oscillates around the local minima, yet the training progresses  along the ``river'' when using the WSD learning rate scheduling \citep{wen2024understanding}.
When Nesterov's lookahead is applied to this training trajectory, the extrapolation is steered toward the overshooting direction, causing the lookahead position to move into regions with higher objective values, see Figure \ref{fig:river_valley} (left). {This phenomenon is further validated in Figure~\ref{fig:nanogpt_lookahead}, where we measure the validation loss at the lookahead positions during NanoGPT training for different lookahead step sizes $\beta$. We observe that increasing the lookahead step size leads to a sharp increase in the validation loss evaluated at the extrapolated positions. This suggests that short-horizon lookahead directions become unreliable in deep learning optimization and may produce extrapolated solutions that are unsuitable for constructing the next optimization step.}

Inspired by the above illustrations, we postulate that an appropriate way to accelerate deep learning model training is to \emph{smooth} the lookahead directions through averaging the past lookahead directions. 
 A line of recent works \citep{zhang2019lookahead,douillard2023diloco,zuo2024nala,kallusky2025snoo} partially addressed this issue by 
 incorporating an inner loop to accumulate (and thus average) the sequence of base optimizer updates and effectively performs long-horizon lookahead.
 These algorithms have demonstrated an empirical acceleration by adopting long-horizon lookahead. 
 However, it is not fully understood how does the mechanism behind long-horizon lookahead enables acceleration.
 We draw insight from the perspective of low-pass filtering and provide a new understanding about long-horizon lookahead as a low-pass filter of the iterate updates.

This work aims at designing a stablized lookahead algorithm that adapts to the varying training trajectory by the use of exponential moving average (EMA) recursion, i.e., in the form of ${\bf m}^t = \gamma {\bf m}^{t-1} + (1-\gamma)({\bf x}^{t} - {\bf x}^{t-1} )$, $\gamma \in (0,1)$, to smooth the lookahead updates. As seen in Figure \ref{fig:river_valley} (right), the EMA recursion finds the direction with less fluctuations; and the {\color{green!50!black}dashed green} curves in Figure \ref{fig:nanogpt_lookahead} shows that {the lookahead position generated by the {\algname} algorithm does not suffer from the same performance degradation as the usual lookahead direction in {\color{red} solid red} curves, therefore can be {\it reliably} used to build the subsequent steps}.  
As a comparison, we summarize the existing lookahead algorithms according to their lookahead direction, lookahead frequency and adaptivity in Table \ref{tab:compare_algos} (also see Section \ref{sec:understanding} and \ref{sec:opt_long_hori}). These recent approaches either suffer from non-adaptive lookahead or short-horizon oscillation due to their implicit algorithmic structure.

\begin{table}[t]
    \def\arraystretch{0.9}
    \centering
    \scalebox{0.8}{
    \begin{tabular}{l|c|c|c}
    \toprule 
    {\bf Prior Works} & {\bf Lookahead Direction} & \makecell{\bf Lookahead \\ \bf Frequency} & {\bf Adaptivity} \\
    \midrule
    {Nesterov's Accelerated Gradient} {\citep{nesterov1983method}} & $\bx^{t} - \bx^{t-1}$ & $1$ & $\checkmark$  \\
    \midrule
    {Pessimistic Lookahead} {\citep{zhang2019lookahead}} & $-(\bx^{t} - \bx^{t-K})$ & $1/K$ & \ding{55} \\
    \midrule
    {NALA} {\citep{zuo2024nala}} &  $\bx^t - \bx^{t-K}$ & $1/K$ & \ding{55} \\
    \midrule
    {SNOO} {\citep{kallusky2025snoo} (see App. \ref{sec:compare_snoo})} & $\bx^{t} - \bx^{t-K}$ (Implicit) & $1/K$ & \ding{55} \\
    \midrule
    {GPA} {\citep{defazio2025smoothing} (see App. \ref{sec:compare_gpa})} & \makecell{${\bf z}^{t} - {\bf z}^{t-1}$ \\ $+ \beta({\bf z}^t - \text{EMA}(\{{\bf z}^{r}\}_{r=0}^{t-1}))$} (Implicit) & $1$ & \ding{55} \\[.15cm]
    \midrule
    \cellcolor{black!10} {\bf \algname} {\bf (Ours)} & \cellcolor{black!10} EMA$(\{\bx^{r} - \bx^{r-1}\}_{r=1}^{t})$ & \cellcolor{black!10} $1$ & \cellcolor{black!10}  $\checkmark$ \\
    \bottomrule
    \end{tabular}
    }
    \caption{Comparison between lookahead algorithms on the direction and frequency of lookahead. Adaptivity refers to whether the lookahead direction can adapt to changes in the trend of training trajectory, e.g., as illustrated in Figure \ref{fig:U_shaped_river_valley}.
    }\vspace{-1.75\baselineskip}
    \label{tab:compare_algos}
\end{table}

\vspace{-0.5\baselineskip}
\paragraph{Our Contributions} The contribution of this paper is summarized below.
\vspace{-0.5\baselineskip}

\begin{itemize}[leftmargin=*]
\item We introduce a trajectory-based perspective of lookahead acceleration for deep learning optimization. We show that classical short-horizon lookahead becomes unstable in deep learning optimization due to oscillatory high-frequency trajectory components, motivating a low-pass filtering view of acceleration. 

\item We propose {\algname}, a lightweight single-loop acceleration method that replaces long-horizon extrapolation with an exponential moving average of optimization trajectories, i.e., for a given base optimizer function ${\cal A}_t: \mathbb{R}^d \rightarrow \mathbb{R}^d$, we perform
\begin{equation}
\begin{aligned}
\bx^{t+1} &= {\cal A}_t(\bx^t + \beta_t \bm^t), \quad 
\bm^{t+1} =  \gamma \bm^t + (1-\gamma) (\bx^{t+1} -  \bx^t),
\end{aligned}
\end{equation}
see Algorithm \ref{alg:ema_nesterov} for details.
{\algname} exhibits a low-passed lookahead direction that can adapt to a varying trend in the training trajectory.

\item We prove that {\algname} retains accelerated convergence rates in convex optimization. In particular, we establish accelerated rates of $O((1-\sqrt{(1-\gamma)/\kappa})^T)$ for strongly convex objectives with condition number $\kappa > 0$ and $O(1/T^2)$ for convex objectives.

\item We demonstrate that {\algname} acts as a generic acceleration wrapper for modern neural network optimizers, consistently improving Adam, SOAP, Muon, and NorMuon on NanoGPT and Llama pre-training benchmarks. For example, in the NanoGPT benchmark\footnote{From \url{https://github.com/KellerJordan/modded-nanogpt}.}, it improves the fine-tuned Muon optimizer on NanoGPT (162M) by achieving a 6\% acceleration (for reaching the benchmark target of 3.28 validation loss).
Our method outperforms existing lookahead algorithms under a comprehensive hyperparameter sweep for all baselines (see Figure \ref{fig:nanogpt_ema_pessimistic_gpa_snoo}). 
\end{itemize}

More broadly, {\algname} can be viewed as a lightweight and optimizer-agnostic acceleration mechanism that can be combined with a broad class of modern neural network optimizers. Rather than redesigning optimizer-specific update rules, {\algname} operates directly at the trajectory level through stabilized lookahead extrapolation, making it naturally compatible with existing preconditioning, normalization, and momentum techniques.

\vspace{-0.5\baselineskip}
\section{Understanding Lookahead Acceleration in Deep Learning} \label{sec:understanding}
\vspace{-0.5\baselineskip}
This section studies the recent prior arts on lookahead acceleration and compares them with the classical Nesterov's lookahead algorithm from a trajectory perspective. 
We then identify key insights in the prior arts and suggest how they may inspire new algorithms. 

To fix notations, we consider the stochastic optimization problem: for random sample $\xi$ drawn from some distribution ${\cal D}$,
\begin{equation} \label{eq:opt}
    f^\star := \textstyle \min_{ \bx \in \mathbb{R}^d }~f( \bx ) := \mathbb{E}_{ \xi \sim {\cal D} } [ f( \bx; \xi ) ],
\end{equation}
where $f( \bx; \xi )$ is differentiable w.r.t.~$\bx$. Here, we may denote ${\cal A}(\bx; \xi, {\bf s})$ as an optimizer function that takes the current model state $\bx$ and returns the next model state, via consuming a sample of stochastic objective function gradient evaluated on $\bx$ with randomness $\xi$, while the optimizer update is conditioned on its internal states ${\bf s}$. For example, ${\cal A}(\bx; \xi, {\bf s}) = \bx - \alpha_t ( \gamma {\bf s} + (1-\gamma) \nabla f(\bx; \xi))$ with learning rate $\alpha_t > 0$ produces the next model state using momentum SGD under gradient momentum rate $\gamma \in [0, 1)$ and momentum buffer ${\bf s} \in \mathbb{R}^d$. For simplicity, in the rest of this paper, we shall use the notation ${\cal A}_t({\bf x}) := {\cal A}({\bf x}; \xi^t, {\bf s}^t)$.

We begin our endeavor through studying the lookahead algorithm in 
\cite{nichol2018first,zhang2019lookahead}. With a lookahead step size $\beta > 0$ and an optimizer ${\cal A}_t: \mathbb{R}^d \to \mathbb{R}^d$ (e.g., vanilla/momentum SGD, Muon, etc.) for \eqref{eq:opt}, they proposed
\begin{equation} \label{eq:pes_lookahead} 
    \begin{aligned}
    \textbf{for every}&~ t=nK, n = 0, ..., \lfloor T/K \rfloor,\\
        \tilde{\bx}^{t+K} &= {\cal A}_{t+K-1}(~\cdots ~{\cal A}_{t+1}( {\cal A}_t(\bx^t) ) ) , \quad \bx^{t+K} = \bx^t + \beta (\tilde{\bx}^{t+K} - \bx^t) .
    \end{aligned}
\end{equation}
For every $K$ steps of optimizer update ${\cal A}_t$, the accumulated update $\tilde{\bx}^{t+K} - \bx^t$ is applied with a small step size $\beta \leq 1$, which means that the algorithm's solution only interpolates between $\tilde{\bx}^{t+K}$ and $\bx^t$. 
{The key property of \eqref{eq:pes_lookahead} is that the long-horizon update direction $\tilde{\bx}_{t+K}-\bx_t$ averages optimization trajectories across multiple iterations, thereby suppressing high-frequency oscillations in stochastic optimization. As a result, the extrapolation direction becomes substantially more stable than the one-step lookahead direction used in classical Nesterov acceleration (similar to the case of Figure \ref{fig:river_valley} (right)).}
Importantly, \citep[Proposition 2]{zhang2019lookahead} demonstrated that the above scheme achieves a ${\cal O}(\beta)$-factor variance reduction. That being said, as demonstrated in \citep[Appendix A]{zhang2019lookahead}, \eqref{eq:pes_lookahead} only exhibits a slightly improved empirical convergence speed over vanilla SGD in the quadratic setting,
suggesting an `unaccelerated' rate of $f( \bx^T) - f^\star = {\cal O}( (1-1/\kappa)^T )$, where $\kappa$ is the condition number of $f (\bx)$.

For further insights, we compare \eqref{eq:pes_lookahead} to the classical Nesterov's accelerated gradient method by reducing \eqref{eq:pes_lookahead} to the special case with $K=1$. Notice that in this case, \eqref{eq:pes_lookahead} is equivalent to:
\begin{equation} \label{eq:hinton_k_eq_1}
\begin{aligned}
    \bx^{t+1} &= \bx^t + \beta ({\cal A}_{t}(\bx^t) - \bx^t) 
    = 
    \underbrace{{\cal A}_{t}(\bx^t) 
    + (\beta - 1) ({\cal A}_{t}(\bx^t) - \bx^t)}_{\text{lookahead position}} .
\end{aligned}
\end{equation}
On the other hand, Nesterov's lookahead algorithm \citep{nesterov1983method} can be written as:
\begin{equation} \label{eq:nesterov_k_eq_1}
\bx^{t+1} = {\cal A}_{t}(~ \underbrace{\bx^t + \beta' (\bx^t - \bx^{t-1})}_{\text{lookahead position}}~),\quad\beta' > 0.
\end{equation}
It is known that \eqref{eq:nesterov_k_eq_1} achieves an accelerated convergence rate: with an appropriate $\beta'$ and ${\cal A}_t$ utilizing deterministic gradient, one has $f( \bx^T ) - f^\star = {\cal O}( (1 - \sqrt{1/\kappa} )^T )$ for strongly convex $f$.

Comparing \eqref{eq:hinton_k_eq_1} and \eqref{eq:nesterov_k_eq_1} on their corresponding lookahead position, 
we note that \eqref{eq:hinton_k_eq_1} can be regarded as a \emph{pessimistic} lookahead algorithm since it uses the reversed Nesterov's lookahead direction when $\beta-1$ becomes negative.
This also implies that \eqref{eq:hinton_k_eq_1} may not be able to achieve acceleration when using a conservative lookahead step size $\beta \leq 1$. Meanwhile, Nesterov's lookahead \eqref{eq:nesterov_k_eq_1} takes an \emph{optimistic} lookahead direction as it fully trusts the update direction of past iteration with any $\beta' > 0$.

\vspace{-0.5\baselineskip}
\section{Stabilized Nesterov's Lookahead} \label{sec:opt_long_hori}
\vspace{-0.5\baselineskip}
\begin{figure}[t]
    \centering
    \includegraphics[width=0.9\linewidth]{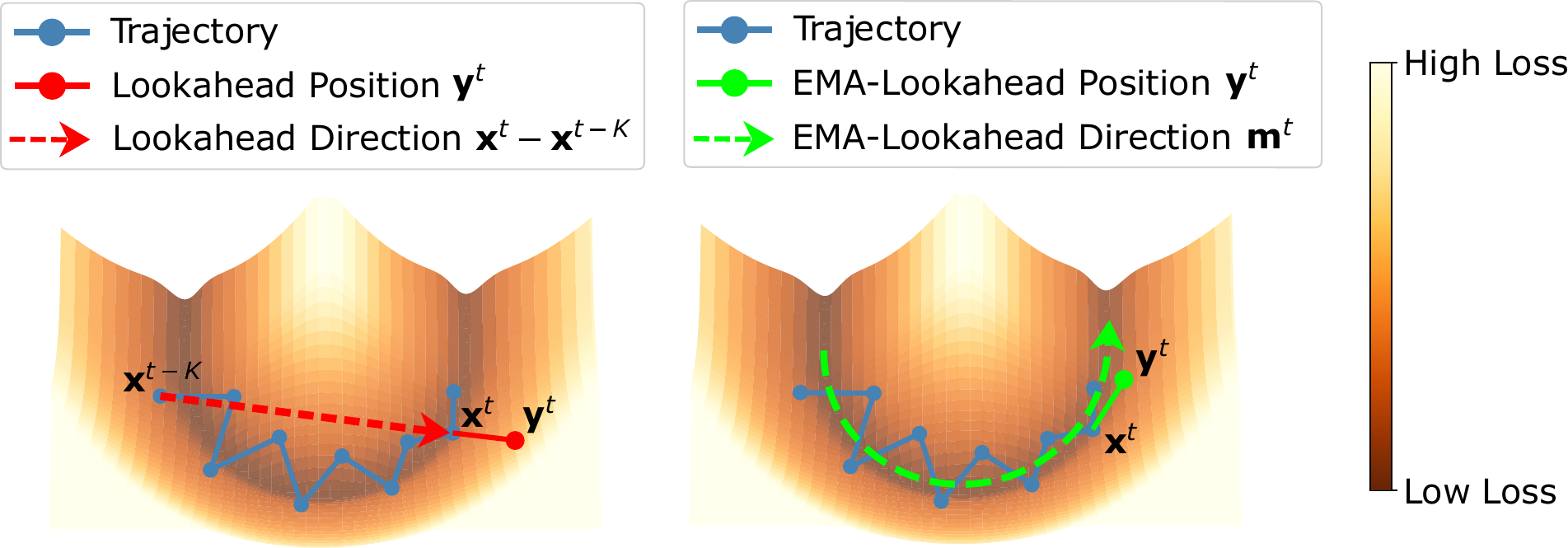}
    \caption{A hypothetical illustration of the training trajectory $(x_1, x_2, f(x_1, x_2)) \in \mathbb{R}^3$, assuming  a U-shaped river valley loss landscape and a training trajectory under the WSD learning rate scheduling \citep{wen2024understanding}. We compare the function value of the long-horizon lookahead position at (Left) $\by^t = \bx^t + \beta \cdot (\bx^{t} - \bx^{t-K})$ which arrives at a high loss position and (Right) $\by^t = \bx^t + \beta \cdot {\bf m}^t$ which progresses towards the trend of training trajectory where ${\bf m}^t = \gamma {\bf m}^{t-1} + (1-\gamma) (\bx^t - \bx^{t-1})$ for $\gamma \in (0,1)$. }
    \label{fig:U_shaped_river_valley}
\end{figure}

We will develop {\algname} using the insights from the trajectory perspective, and to present the convergence properties of {\algname}. The discussion from the previous section motivates us to look at an optimistic lookahead algorithm with $K$-step long-horizon lookahead:
\begin{equation}\label{eq:nesterov_K}
    \begin{aligned}
    \textbf{for every}&~ t=nK, n = 0, ..., \lfloor T/K \rfloor,\\
        \bx^{t+K} &= {\cal A}_{t+K-1}(~\cdots ~{\cal A}_{t+1}( {\cal A}_t(\bx^t + \beta' {\bf b}^t) ) ), \quad {\bf b}^{t+K} = \bx^{t+K} - \bx^{t} .
    \end{aligned}
\end{equation}
The same algorithm structure is used explicitly in NALA \citep{zuo2024nala} or implicitly in SNOO \citep{kallusky2025snoo}.
Particularly, \cite{zuo2024nala} attempted to combine the advantages of Nesterov's accelerated gradient method and long-horizon lookahead, yielding an algorithm similar to \eqref{eq:nesterov_K} where ${\cal A}_t$ is gradient descent for $t ~{\rm mod} ~K = 0$ and ${\cal A}_t$ is a pre-conditioned stochastic algorithm for $t ~{\rm mod} ~K\neq 0$. Later, \cite{kallusky2025snoo} treated the long-horizon lookahead direction as a pseudo-gradient for Nesterov's momentum algorithm \citep{sutskever2013importance}, yielding an algorithm that is fundamentally similar to \eqref{eq:nesterov_K} as discussed in Appendix \ref{sec:compare_snoo}. \cite{defazio2025smoothing} developed a single-loop algorithm under the primal-averaging framework which implicitly approximated a mixture of short-horizon and long-horizon lookahead as discussed in Section \ref{sec:compare_gpa}, where such mixture implies unstable oscillation from short-horizon lookahead. We summarize the above algorithms from the lookahead perspective in Table \ref{tab:compare_algos}.

To fully understand the long-horizon lookahead direction ${\bf b}^t$ in \eqref{eq:nesterov_K}, we rewrite ${\bf b}^t$ in terms of one-step iterate update $\Delta \bx^t := \bx^t - \bx^{t-1}$ as
\begin{equation} \label{eq:low_pass_lookahead}
\textstyle {\bf b}^t = \bx^{t} - \bx^{t-K} = K \cdot \frac{1}{K}\sum_{i=t-K+1}^{t} \Delta \bx^i,~t \geq K. 
\end{equation}
As seen, ${\bf b}^t$ is the unweighted combination of the past $K$ iterations of (one-step) lookahead directions, i.e., it is the output of an order-$K$ moving average \emph{low-pass filter}
over the sequence $\{ \Delta \bx^i \}_{i \geq 1}$. The low-pass filter suppresses high-frequency fluctuations in consecutive iterate updates. This aligns with our prior understanding that long-horizon lookahead can reduce fluctuation in the lookahead updates. However, we argue that such unweighted combination induces a lookahead direction that is not adaptive to the local geometry. In particular, when the trend of training trajectory evolves over time, the unweighted average \eqref{eq:low_pass_lookahead} with large $K$ can only capture a global trend, which is not suitable for local extrapolation, as illustrated in the left figure of Figure \ref{fig:U_shaped_river_valley}.

\vspace{-0.5\baselineskip}
\subsection{Exponential Moving Average Lookahead}
\vspace{-0.5\baselineskip}

Alternative to long-horizon lookahead, we replace the lookahead direction with a low-pass filter that is adaptive to the recent signals --- the exponential moving average (EMA) filter: for $t \geq 1$ and let $\gamma \in [0,1)$,
\begin{equation} \label{eq:ema_lookahead}
\textstyle \bm^{t+1} = \gamma \bm^t + (1-\gamma) \Delta \bx^{t+1} =: {\rm EMA}_\gamma ( \{ \Delta \bx^i \}_{i=1}^{t+1} ).
\end{equation}
In particular, one has ${\rm EMA}_\gamma ( \{ \Delta \bx^i \}_{i=1}^{t+1} ) = (1-\gamma) \sum_{i=1}^{t+1} \gamma^{t+1-i} \Delta \bx^i$.
Notice that \eqref{eq:ema_lookahead} is a low-pass filter with the geometric coefficients $\{ \gamma^{i} \}_{i \geq 1}$ that is anticipated to filter out the high-frequency fluctuations while putting larger weights on recent lookahead signals $\Delta {\bf x}^i$, i.e., it aligns more with the recent trend of $\Delta \bx^i$ as illustrated in the right figure of Figure \ref{fig:U_shaped_river_valley}. The low-pass behavior of EMA filter can be analyzed by its transfer function $H(z) = \frac{1-\gamma}{1-\gamma z^{-1}}$ on the input signal $z=\exp(i\omega)$ of frequency $\omega \in [0, \pi]$, where the output energy $|H(\exp(i\omega))|$ is decreasing as $\omega$ increases from $0$ (low frequency) to $\pi$ (high frequency). 
We replace the lookahead direction of Nesterov's algorithm by ${\bf m}^t$ and propose {\algname} in Algorithm \ref{alg:ema_nesterov}.

\begin{algorithm}
\caption{{\tt EMA-Nesterov} Algorithm}
\label{alg:ema_nesterov}
\begin{algorithmic}[1]
\STATE {\bfseries Input:} Initial model ${\bf x}^0$, lookahead step size $\beta_t \ge 0$, exponential moving average rate $\gamma \in [0, 1)$, base optimizer ${\cal A}_t(\cdot )$, iterations $T > 0$.
\STATE {\bfseries Initialization:} $\bm^0 = {\bf 0}$.
\FOR{$t=0, ..., T-1$}
\STATE \label{alg_step:lookahead_opt} $\bx^{t+1} = {\cal A}_t(\bx^t + \beta_t \bm^t)$
\STATE \label{alg_step:ema_lookahead} $\bm^{t+1} =  \gamma \bm^t + (1-\gamma) (\bx^{t+1} -  \bx^t) $

\ENDFOR
\STATE {\bfseries Output:} Last iterate solution $\bx^T$.
\end{algorithmic}
\end{algorithm}

Line \ref{alg_step:lookahead_opt} of Algorithm \ref{alg:ema_nesterov} performs an optimization step from the lookahead position $\bx^t + \beta_t {\bf m}^t$, using the EMA lookahead direction ${\bf m}^t$. With $\gamma > 0$, Algorithm \ref{alg:ema_nesterov} accumulates a stabilized lookahead direction in ${\bf m}^t$ that effectively follows the trend of optimization trajectory, without being perturbed by the oscillation that enters the filter as high frequency noise.

\textbf{Comparison to Existing Algorithms.} 
We can directly compare the differences between existing lookahead algorithms by reducing them into the lookahead form. For example, SNOO \citep{kallusky2025snoo} can be reduced to the $K$-step long-horizon lookahead in \eqref{eq:nesterov_K} (see Appendix \ref{sec:compare_snoo} for the derivation) and GPA \citep{defazio2025smoothing} can be reduced to the following lookahead algorithm (see Appendix \ref{sec:compare_gpa} for the derivation):
\begin{equation} 
    \begin{aligned}
    \textbf{for every}&~ t=0, ..., T-1,\\
        {\bf z}^{t+1} &= {\bf z}^t - \gamma_p \nabla f({\bf y}^t; \xi^t), \quad {\bf x}^{t+1} = \beta {\bf x}^t + ( 1 - \beta) {\bf z}^{t+1}, \\
        {\bf y}^{t+1} &= {\bf y}^{t} + (1- \beta)({\bf z}^{t+1} - {\bf z}^t) + \beta(1-\beta)({\bf z}^{t+1} - {\bf x}^t).
    \end{aligned}
\end{equation}
The lookahead direction of GPA is determined by ${\bf z}^{t+1} - {\bf z}^t$ and ${\bf z}^{t+1} - \bx^t$, where the former is vulnerable to high-frequency fluctuations while the latter is effectively a long-horizon lookahead with $\bx^t$ tracking the past trajectory of ${\bf z}^t$. A direct comparison on the numerical performance is provided in Figure \ref{fig:nanogpt_ema_pessimistic_gpa_snoo}, where our algorithm further improved GPA by 0.2 validation perplexity on NanoGPT. 
Theoretically, the convergence analysis of \cite{defazio2025smoothing} does not guarantee accelerated convergence rate. In contrary, our Algorithm \ref{alg:ema_nesterov} adopts a matching design with Nesterov's accelerated algorithm which enables us to carry out convergence analysis with accelerated rate for convex optimization.

\vspace{-0.5\baselineskip}
\subsection{Convergence Analysis} \label{sec:convergence}
\vspace{-0.5\baselineskip}
In this subsection, we show that due to the optimistic design, {\algname} attains the same accelerated convergence as the classical Nesterov's accelerated gradient algorithm when specialized to \eqref{eq:opt} with (strongly) convex objective function and the optimizer utilizes exact gradients of $f(\bx)$.
We focus on the special case of Algorithm \ref{alg:ema_nesterov} with the optimizer specified as the vanilla gradient descent: 
\begin{equation} \label{eq:deterministic_sp_case}
    {\cal A}( \bx; \xi; {\bf s} ) = \bx - \alpha \nabla f( \bx ) .
\end{equation}
We notice that Algorithm \ref{alg:ema_nesterov} with \eqref{eq:deterministic_sp_case} and EMA rate $\gamma = 0$ reduces to the classical Nesterov's accelerated gradient algorithm; yet when $\gamma > 0$, the two algorithms exhibit different lookahead behavior. Despite their similar algorithmic structure, we notice that extending the analysis in \citep{nesterov1983method} to Algorithm \ref{alg:ema_nesterov} is highly non-trivial, especially for achieving the accelerated convergence rate(s). 
In this section, we proceed with the convergence guarantee of the above algorithm under a standard $\mu$-strongly convex, $L$-smooth assumption for \eqref{eq:opt} as defined below.
\begin{assumption} \label{assm:strong_cvx}
    For any $L > \mu \ge 0$, we assume the objective function $f$ is $\mu$-strongly convex and has $L$-Lipschitz gradient such that $\| \nabla f(\bx) - \nabla f(\by) \|_2 \leq L \| \bx - \by \|_2~\forall \bx, \by \in \mathbb{R}^d$. We define $\kappa := L / \mu$ as the condition number of $f$.
\end{assumption}

We remark that as the deep learning training problems are typically non-convex, the convexity assumption above may not be satisfied in practice. Nonetheless, they provide insights and theoretical support on the advantages of {\algname}, and the analysis can be of independent interests for studying perturbed Nesterov's accelerated gradient.
Our results are further split into two cases. 
\textbf{Strongly Convex $f(\bx)$.}
When $\mu > 0$, we observe that
\begin{theorem}
    Under Assumption \ref{assm:strong_cvx}, by choosing $\alpha = 1/L$ and $\beta_t = \frac{\left(\sqrt{1 - \gamma} - \sqrt{1/\kappa}\right)^2}{(1-\gamma)(1-1/\kappa)}$ for all $t=0, ..., T-1$, the solution of  Algorithm \ref{alg:ema_nesterov} with \eqref{eq:deterministic_sp_case} satisfies
    \begin{equation}
        f(\bx^T) - f(\bx^\star) = \mathcal{O}\left((1 - \sqrt{(1-\gamma) / \kappa})^T\right) \quad \text{for any $\gamma \in [0, 1-1/\kappa)$, $T > 0$.}
    \end{equation}
\end{theorem}
See Appendix \ref{app:proof_strong_cvx} for the proof. The above guarantee reduces to the classical accelerated rate of $(1-\sqrt{1/\kappa})^T$ of Nesterov's algorithm when $\gamma = 0$. Our analysis generalizes to {\algname} when $\gamma > 0$ and the acceleration upon the gradient descent rate of $(1-1/\kappa)^T$ happens when
\begin{equation}
    \sqrt{(1-\gamma)/\kappa} > 1/\kappa \Leftrightarrow \gamma < 1- 1/\kappa.
\end{equation}

\textbf{Non-strongly Convex $f(\bx)$.} When $\mu = 0$, we observe that 
\begin{theorem} \label{thm:cvx}
    Under Assumption \ref{assm:strong_cvx} with $\mu = 0$, by choosing $\alpha = 1/L$ and $\beta_t = \frac{c_t - \gamma c_{t+1}}{1 + (1-\gamma) c_{t+1}}$ for $c_t = 1 + \frac{\gamma}{4(1-\gamma)} + \frac{t}{4}$, the solution of Algorithm \ref{alg:ema_nesterov} with \eqref{eq:deterministic_sp_case} satisfies
    \begin{equation}
        f(\bx^T) - f(\bx^\star) = {\cal O}\left(L / ( {(1-\gamma)^3T^2} ) \right) \quad \text{for any $\gamma \in [0, 1)$, $T > 0$.}
    \end{equation} 
\end{theorem}
See Appendix \ref{app:proof_cvx} for the proof. Unlike the strongly convex case, Theorem \ref{thm:cvx} guarantees an accelerated convergence rate of $\mathcal{O}(1/T^2)$ for all $\gamma \in [0, 1)$, i.e., it converges faster than gradient descent of rate $\mathcal{O}(1/T)$ on convex objective function for large enough $T = \Omega(1/(1-\gamma)^3)$.

\vspace{-0.5\baselineskip}
\subsection{Practical Implementation of Lookahead Step Size Scheduling}
\vspace{-0.5\baselineskip} \label{sec:beta_schedule}
It is generally understood that the early stage of deep learning optimization is unstable.
As a remedy, we will schedule the lookahead step size $\beta_t$ by three stages: 1) the warm-up stage where $\beta_t = 0$ for $t \leq T_w$; 2) the lookahead stage where $\beta_t = \beta \cdot \alpha_t / (\max_k \alpha_k)$ for $T_w < t \leq T_r$ such that the lookahead step size relatively follows $\alpha_t / (\max_k \alpha_k)$ the learning rate schedule; and 3) the early rest stage where $\beta_t = 0$ for $t > T_r$. The early rest stage is crucial to allow stable convergence to sharp local minima during the learning rate decay phase, as the EMA lookahead direction cannot quickly adapt to the local geometry near sharp minima. This is further validated by the observation in Figure \ref{fig:nanogpt_lookahead_convergence}, where the EMA lookahead direction leads to sharp increase in loss in later stage of the training. In our experiments, the scheduling of $\beta_t$ ($T_w$ and $T_r$) roughly follows the learning rate schedule as the learning rate schedule implicitly controls when stable training occurs. Details of the actual schedule used in our experiments can be found in Appendix \ref{app:exp}.

\begin{figure}[t]
    \centering
    \includegraphics[width=0.325\linewidth]{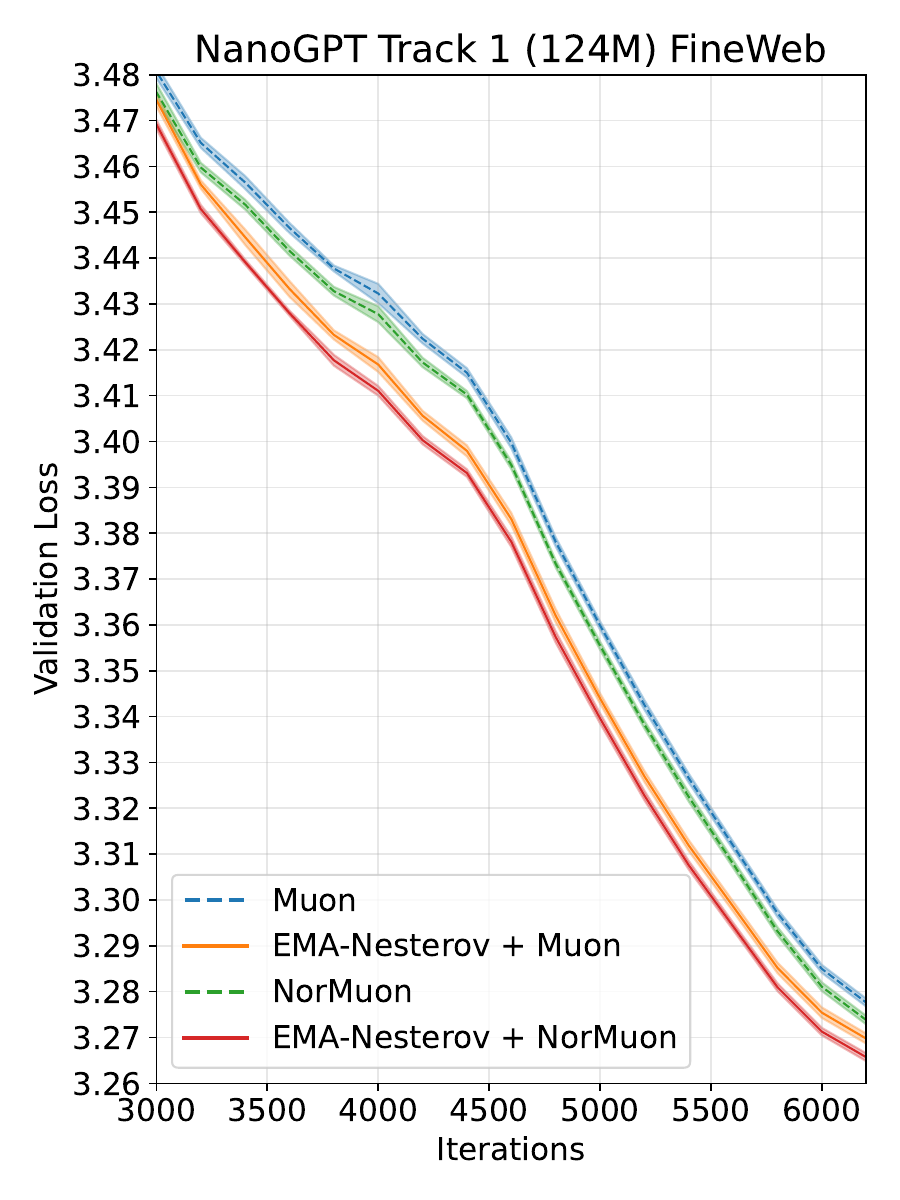}
    \includegraphics[width=0.325\linewidth]{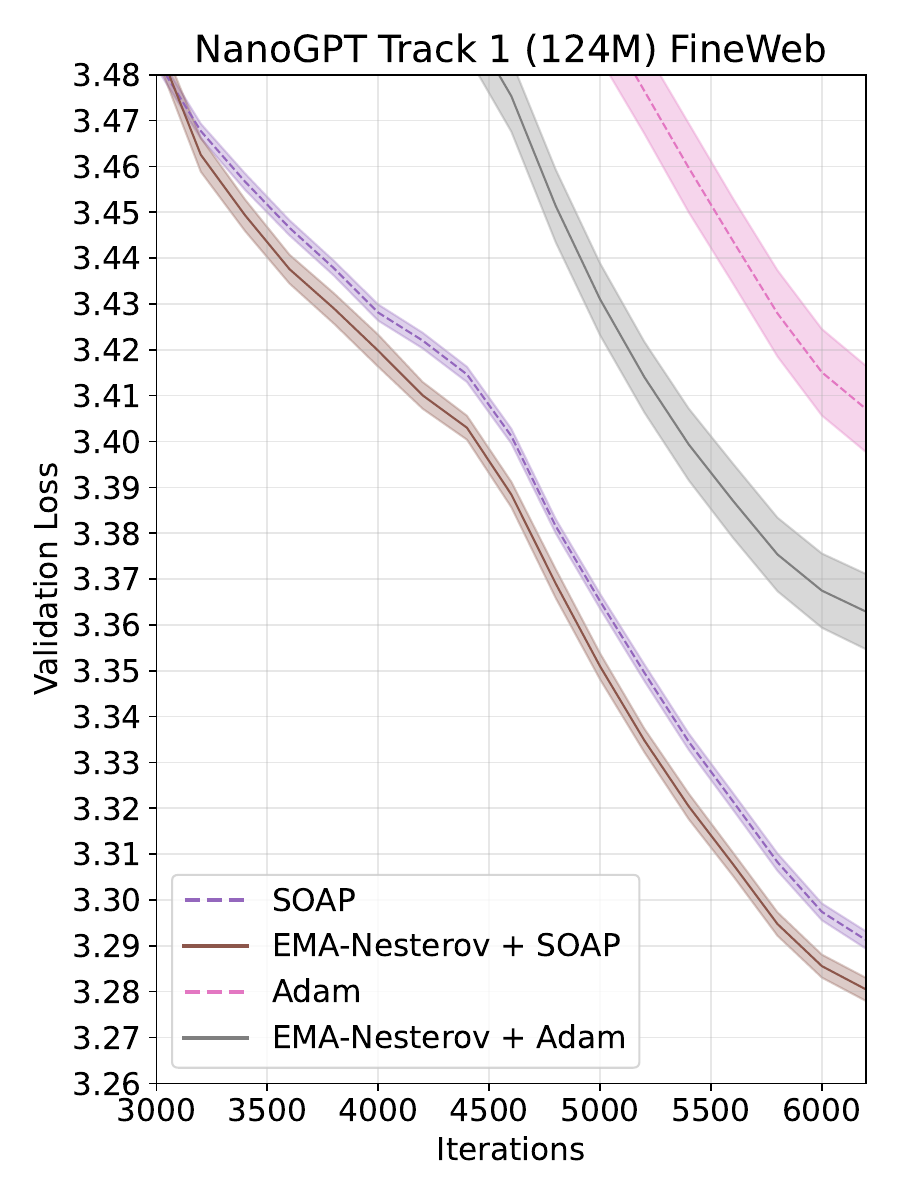}
    \includegraphics[width=0.325\linewidth]{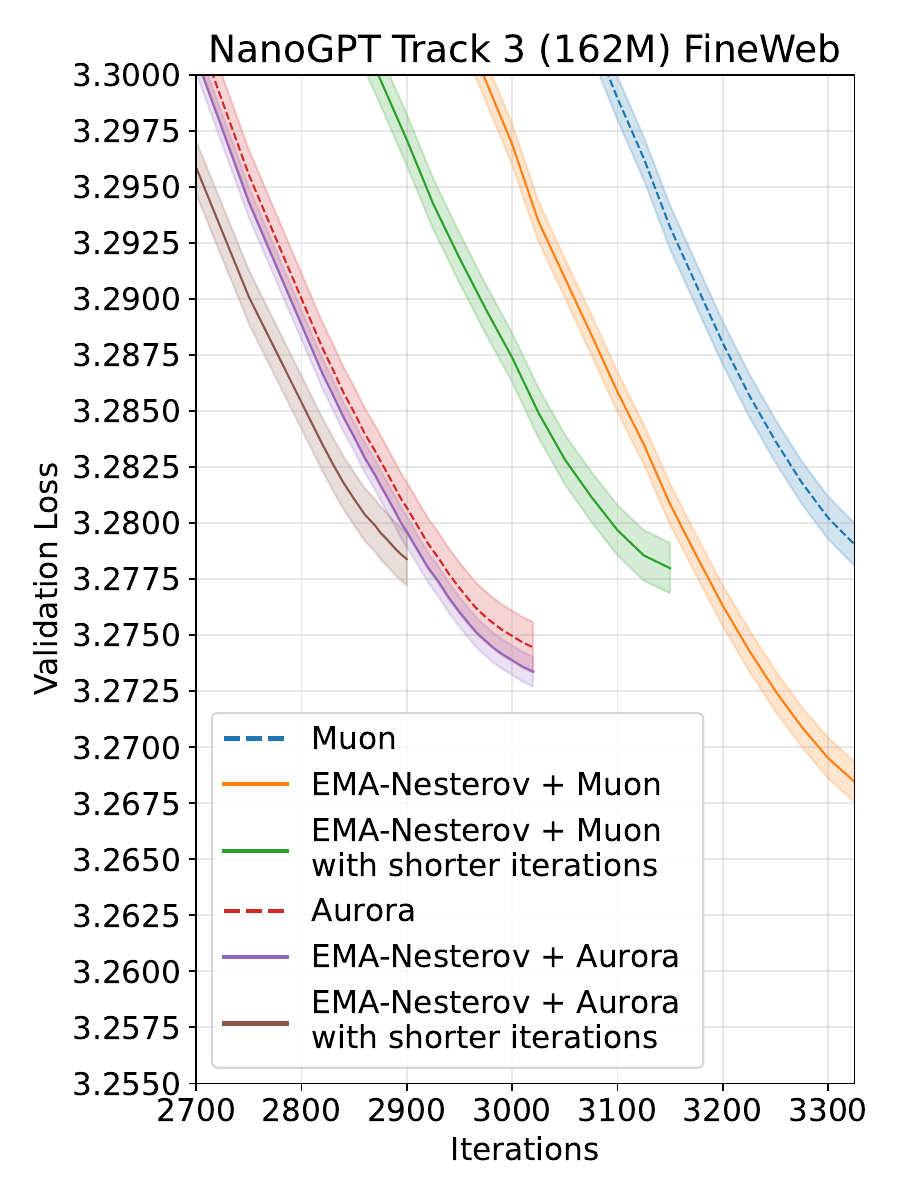}\vspace{-.2cm}
    \caption{Training {\algname} with different base optimizers (Muon, NorMuon, SOAP and Adam) on NanoGPT Track 1 (124M) benchmark (Left, Middle) and with SOTA base optimizers (as of 05/18/2026) on NanoGPT Track 3 (162M) (Right). Results are reported by mean-variance plots of $\pm 1$ standard deviation on multiple seeded runs for each algorithm.}\vspace{-.4cm}
    \label{fig:nanogpt_ema_base_opts}
\end{figure}

\begin{figure}[t]
    \centering
    \includegraphics[width=0.99\linewidth]{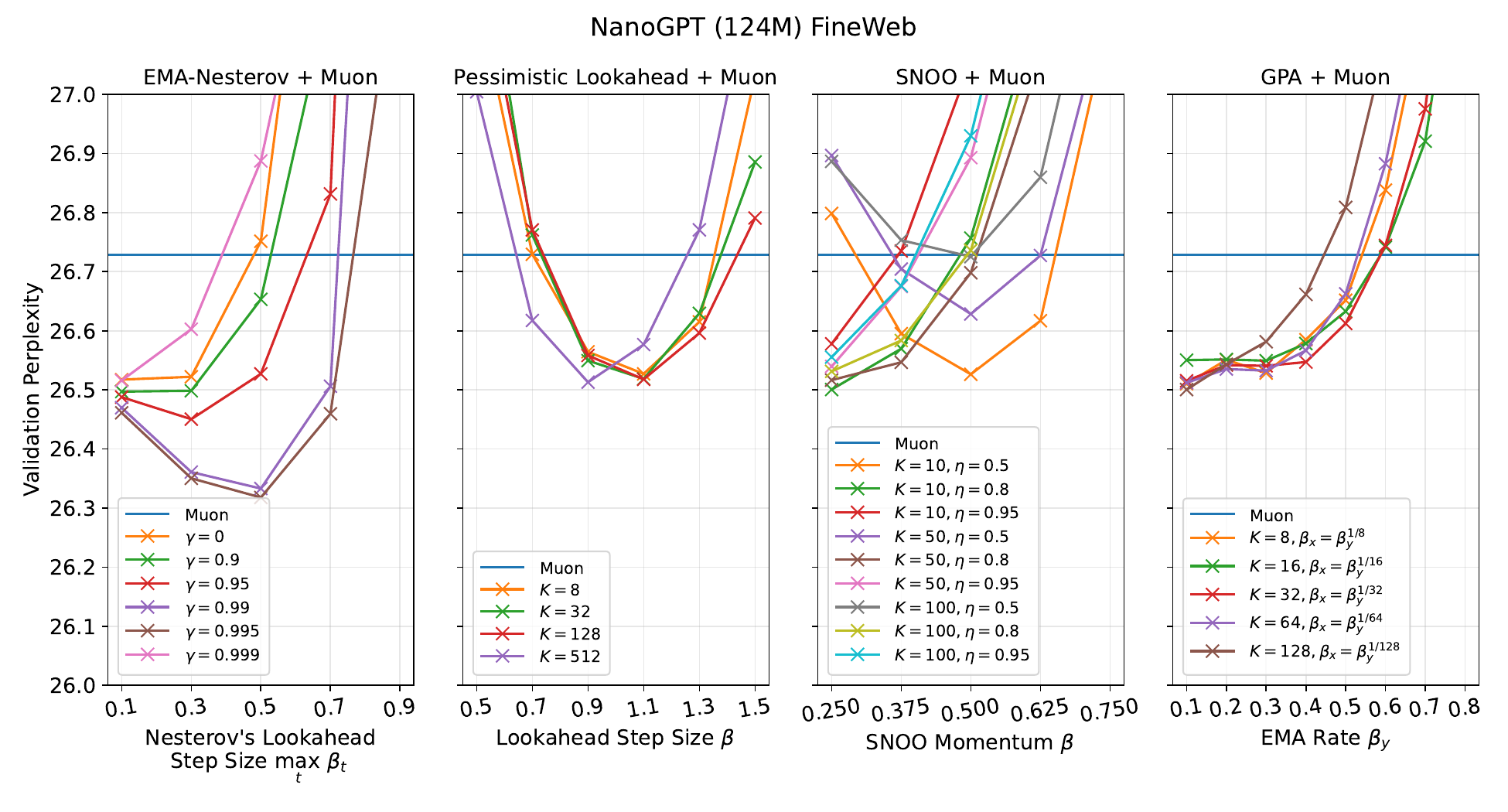}\vspace{-.3cm}
    \caption{Comparison on accelerating Muon base optimizer using {\algname}, pessimistic lookahead, GPA and SNOO on NanoGPT (124M) benchmark. The same vertical axis spans across four plots horizontally. Among all hyperparameter configurations, {\algname} + Muon with $\gamma = 0.995$, $\max_t \beta_t = 0.5$ achieves the lowest validation perplexity.}\vspace{-.5cm}
    \label{fig:nanogpt_ema_pessimistic_gpa_snoo}
\end{figure}

\vspace{-0.5\baselineskip}
\section{Experiments} \label{sec:exp}
\vspace{-0.5\baselineskip}
We validate the performance of Algorithm \ref{alg:ema_nesterov} on two language model pre-training settings: {\bf (I)} training the NanoGPT models on FineWeb dataset\footnote{124M model is based on \url{https://github.com/KellerJordan/modded-nanogpt/blob/master/records/track_1_short/2024-10-10_Muon/eb5659d0-fb6a-49e5-a311-f1f89412f726.txt} while 162M model is based on \url{https://github.com/KellerJordan/modded-nanogpt/tree/master/records/track_3_optimization}.} and {\bf (II)} training Llama models on C4-en dataset. Unless otherwise specified, all settings use a training budget of around 20 tokens-per-parameter as suggested by \cite{hoffmann2022training}. The main experiments are performed on 8$\times$ NVIDIA H200 GPUs (141 GB memory). Our algorithm is implemented as a lightweight wrapper and does not impose extra overhead on the per-iteration time, we thus focus on comparing different algorithms' performance after a fixed number of training iterations. The base optimizers' hyperparameters are tuned without {\algname} and are unchanged when adding {\algname} externally. 
Our experiment is open-sourced at \url{https://github.com/OptimAI-Lab/ema-nesterov}.

\textbf{Accelerating Common Optimizers.}
The first experiment aims to present the acceleration effects offered by applying {\algname} on common pre-conditioned optimizers as ${\cal A}_t$ --- including Adam \citep{kingma2014adam}, SOAP \citep{vyas2024soap}, Muon \citep{jordan2024muon} and NorMuon \citep{li2025normuon}. 
In Figure \ref{fig:nanogpt_ema_base_opts}, we apply {\algname} with the best pair of hyperparameters $(\gamma, \max_t \beta_t)$ (selected from Figure \ref{fig:nanogpt_ema_pessimistic_gpa_snoo}) onto different choices of base optimizer ${\cal A}_t$. 
Also from the figure, {\algname} improves the performances of these common optimizers ranging from 0.208 perplexity drop (on Muon) to 1.305 perplexity drop (on Adam).
For the larger model size of NanoGPT, we similarly apply {\algname} onto the SOTA optimization algorithms such as Muon and Aurora \citep{dewulf2026aurora} and observe acceleration upon the base optimizers. Note that to achieve the maximal acceleration for arriving at a target loss value (e.g., the benchmark target of 3.28 validation loss in Figure \ref{fig:nanogpt_ema_base_opts}), we adjust the total iterations accordingly as shown in the right figure of Figure \ref{fig:nanogpt_ema_base_opts}.
These evidences show that {\algname} can offer acceleration to existing fine-tuned optimizers.

\textbf{Comparison with Existing Lookahead Algorithms.}
Our second experiment compares {\algname} to existing baseline lookahead algorithms. For fairness, we also perform a hyperparameter search 
on the NanoGPT setting for pessimistic lookahead \citep{zhang2019lookahead}, SNOO \citep{kallusky2025snoo} and GPA \citep{defazio2025smoothing} according to the corresponding suggested range of hyperparameters. As seen in Figure \ref{fig:nanogpt_ema_pessimistic_gpa_snoo}, {\algname} achieved the largest improvement among the other lookahead algorithms under the Muon base optimizer ${\cal A}_t$.

\begin{figure}[t]
    \centering
    \includegraphics[width=0.9\linewidth]{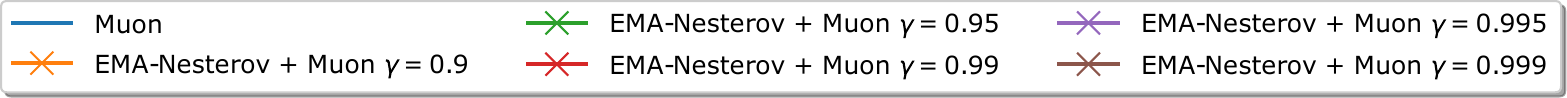}\\
    \includegraphics[width=0.32\linewidth]{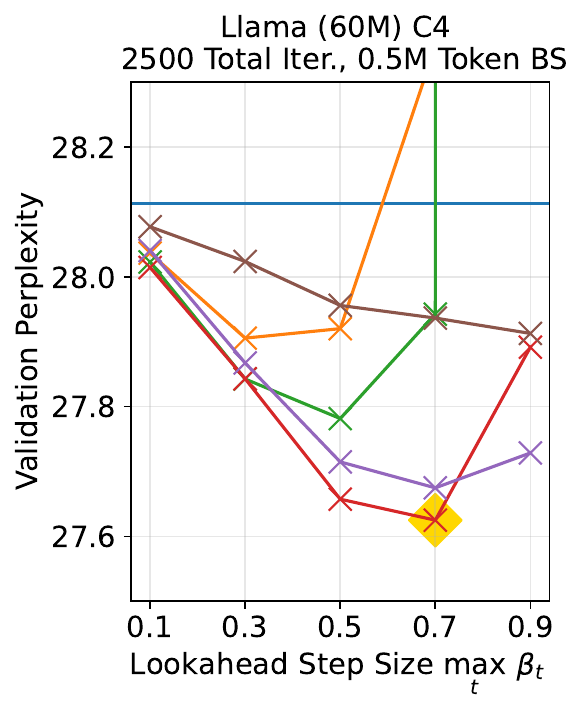}
    \includegraphics[width=0.32\linewidth]{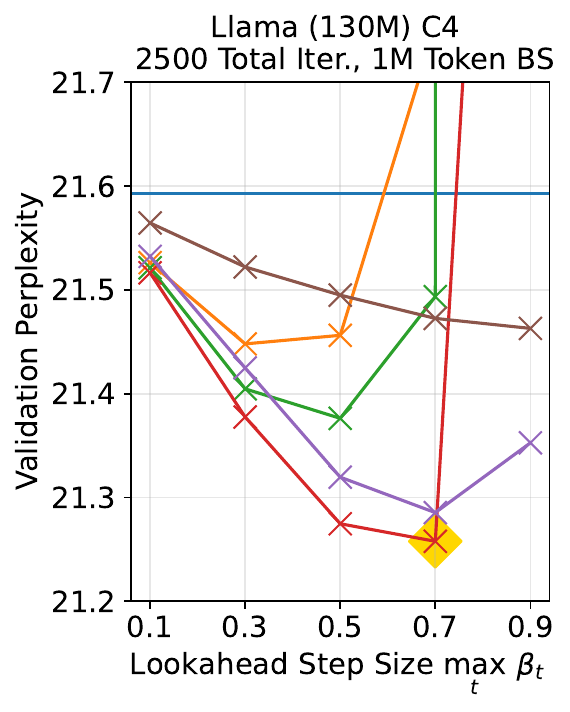}
    \includegraphics[width=0.32\linewidth]{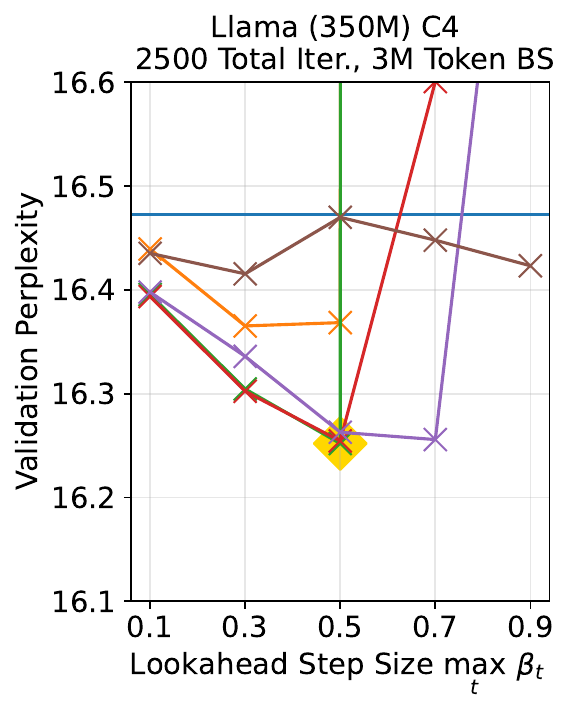}\vspace{-.2cm}
    \caption{Sensitivity of hyperparameters for {\algname} + Muon across different model sizes on the Llama setup. The value of $\max_t \beta_t$ is searched in $\{0.1, 0.3, 0.5, 0.7, 0.9\}$ across the horizontal axis and the value of $\gamma$ is searched in $\{0.9, 0.95, 0.99, 0.995, 0.999\}$ across different colors.}\vspace{-.2cm}
    \label{fig:llama_model_scaling}
\end{figure}
\textbf{Model Scaling.} This experiment investigates how sensitive the performance of {\algname} is on choosing the hyperparameters $\gamma$ and $\max_t \beta_t$ for different model sizes. We perform a set of hyperparameter sweep over 60M, 130M and 350M parameter Llama models on C4 pre-training. The batch sizes are scaled up accordingly to match the total training budget of $20 \times$ tokens-per-parameter. 
From Figure \ref{fig:llama_model_scaling}, when scaling to larger models, the optimal  hyperparameters shift predictably. As the model size increases, the optimal $\gamma$ tends to increase and the optimal $\max_t \beta_t$ tends to decrease. By this set of results, default choices of $\max_t \beta_t = 0.5$, $\gamma = 0.99$ for model size $\leq 350$M and $\gamma = 0.995$ for model size $> 350$M are recommended for large batch setting. In Appendix \ref{app:sensitivity}, we provide more hyperparameter sweeps to illustrate the hyperparameter sensitivity in other settings and our default hyperparameters deliver near-optimal performance. Finally, we scale up to 1B model in the left figure of Table \ref{tab:all_llamas}, and summarizes the last iteration validation perplexity for different Llama models in Table \ref{tab:all_llamas}.

\begin{table}[t]
\begin{minipage}{0.45\textwidth}%
    \includegraphics[width=1\linewidth]{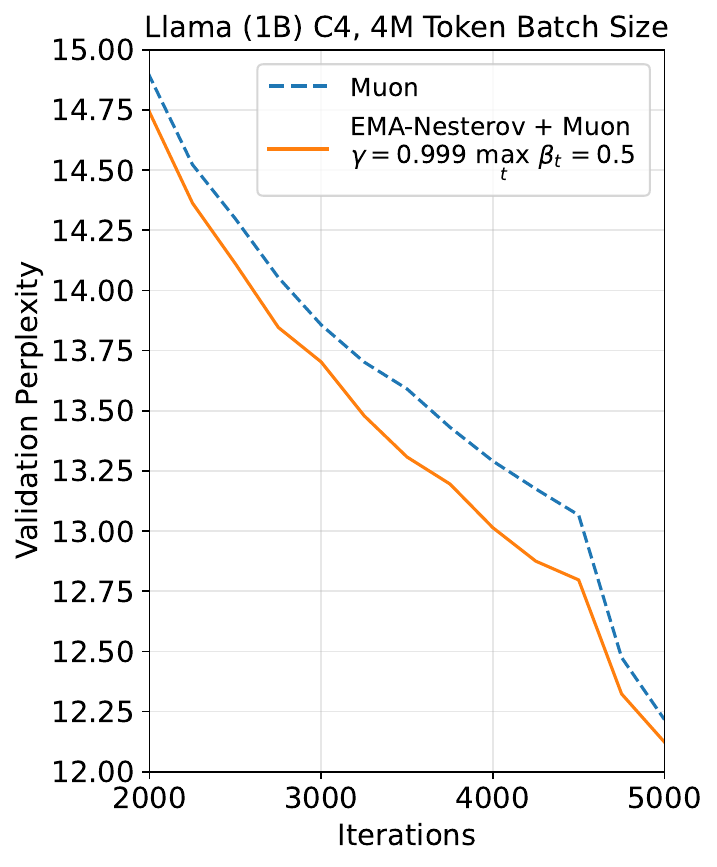}
\end{minipage}
\begin{minipage}{0.5\textwidth}
\label{tab:main_results}
\resizebox{1\textwidth}{!}{%
\begin{tabular}{c|c|cc|cc}
\toprule
 Algorithm &  Val. Ppl. & Iter. & \makecell{Token \\ Batch Size} & $\gamma$ & $\max_t \beta_t$ \\
\midrule
\multicolumn{6}{c}{\textbf{60M Model}} \\
\midrule
Muon & 28.11 & \multirow{2}{*}{2500} & \multirow{2}{*}{0.5M} & - & -\\
{\algname} + Muon & 27.62 &  & & 0.99 & 0.7 \\
\midrule
Muon & 24.09 & \multirow{2}{*}{2500} & \multirow{2}{*}{4M} & - & -\\
{\algname} + Muon & 23.67 &  & & 0.99 & 0.7 \\
\midrule
Muon & 28.21 & \multirow{2}{*}{5000} & \multirow{2}{*}{0.26M} & - & -\\
{\algname} + Muon & 27.27 &  & & 0.995 & 0.7 \\
\midrule
\multicolumn{6}{c}{\textbf{130M Model}} \\
\midrule
Muon & 21.59  & \multirow{2}{*}{2500} & \multirow{2}{*}{1M} & - & -\\
{\algname} + Muon & 21.26  & & & 0.99 & 0.7\\
\midrule
Muon &  18.39 & \multirow{2}{*}{2500} & \multirow{2}{*}{8M} & - & -\\
{\algname} + Muon & 18.11  &  & & 0.99 & 0.7\\
\midrule
Muon & 21.15  & \multirow{2}{*}{5000} & \multirow{2}{*}{0.5M} & - & -\\
{\algname} + Muon &  20.85 &  && 0.995 & 0.5 \\
\midrule
Muon &  17.78 & \multirow{2}{*}{5000} & \multirow{2}{*}{4M} & - & -\\
{\algname} + Muon & 17.59  &  & & 0.995 & 0.5\\
\midrule
\multicolumn{6}{c}{\textbf{350M Model}} \\
\midrule
Muon & 16.12 & \multirow{2}{*}{2500} & \multirow{2}{*}{3M} & - & -\\
{\algname} + Muon & 15.86 &  && 0.99 & 0.5 \\
\midrule
Muon & 14.01 & \multirow{2}{*}{2500} & \multirow{2}{*}{25M} & - & -\\
{\algname} + Muon & 13.82 &  && 0.99 & 0.5 \\
\midrule
\multicolumn{6}{c}{\textbf{1B Model}} \\
\midrule
Muon & 12.22 & \multirow{2}{*}{5000} & \multirow{2}{*}{4M} & - & -\\
{\algname} + Muon & 12.12 &  && 0.999 & 0.5 \\
\bottomrule
\end{tabular}}
\end{minipage}
\caption{
(Left) Training curve of Muon and {\algname} + Muon on Llama 1B for 5000 iterations. (Right) Validation perplexities of different sizes of Llama models. 
}\vspace{-.5cm}
\label{tab:all_llamas}
\end{table}

\begin{figure}[t]
    \centering
    \includegraphics[width=0.325\linewidth]{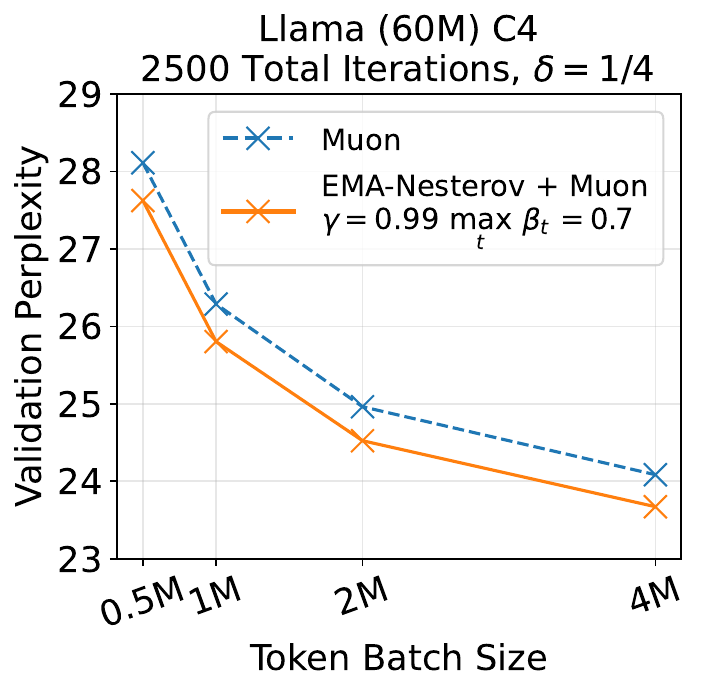}
    \includegraphics[width=0.325\linewidth]{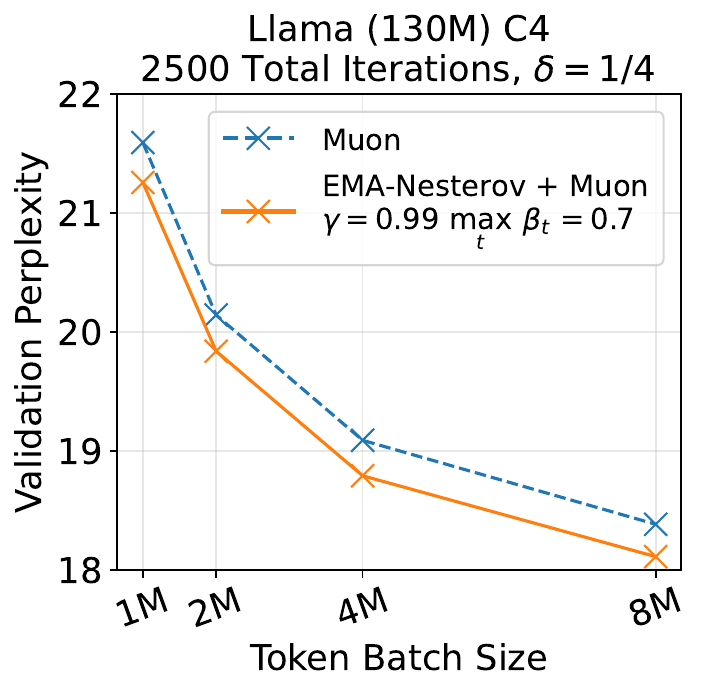}
    \includegraphics[width=0.325\linewidth]{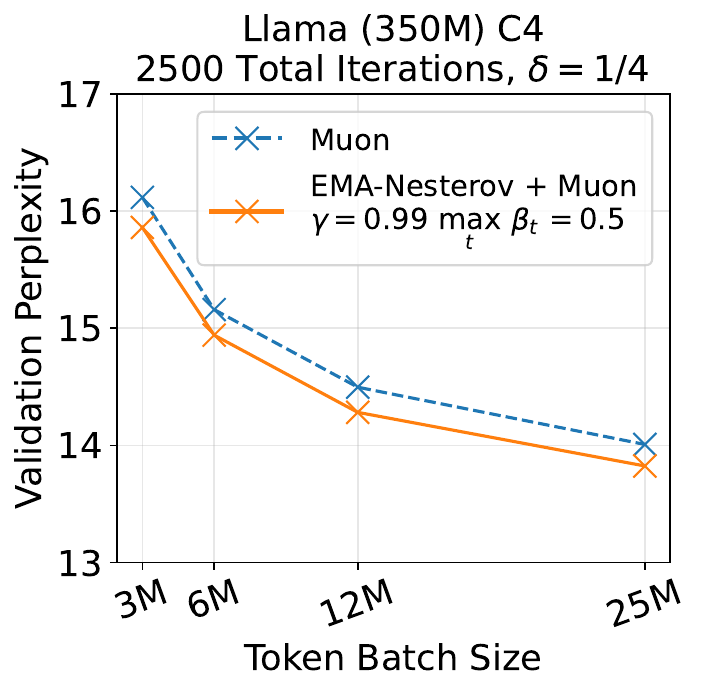}
    \vspace{-.2cm}
    \caption{Data scaling to total training budgets of $\ge \{20\times, 40\times, 80\times, 160\times\}$ tokens-per-parameter with increased token batch size 
    and fixed 2500 total iterations. The learning rates are adapted under the power scaling rule: for tuned learning rate $\alpha$ on batch size $B$, the adapted learning rate $\alpha'$ for larger batch size $B' > B$ is $\alpha' = \alpha \cdot (B'/B)^{\delta}$ with the learning rate scaling exponent $\delta=1/4$.}
    \label{fig:llama_data_scaling}
\end{figure}

\textbf{Data / Batch Size Scaling.}
To stress test the performance of {\algname} on more realistic setting, we increase the training budget beyond $20\times$ tokens-per-parameter in Figure \ref{fig:llama_data_scaling}. Under the optimal hyperparameters from Figure \ref{fig:llama_model_scaling}, we observe that {\algname} consistently improved the base optimizer in larger batch size setting, validating the practical advantage of {\algname} for large-scale data-parallel pre-training. We also observe that a $(\cdot)^{1/4}$ learning rate scaling rule delivers better performance than the square root scaling rule (see comparison in Figure \ref{fig:lr_scaling_comp}).

\section{Conclusion}
We proposed a lightweight wrapper algorithm called {\algname} that is motivated by the low-pass filtering of lookahead directions in deep learning setting. {\algname} utilizes a stabilized lookahead direction by filtering the high frequency oscillation in the training trajectory. We show that {\algname} achieved acceleration in practical deep learning problems and theoretical convex problems.

\bibliographystyle{plainnat}
\bibliography{reference.bib}
\newpage
\appendix

\section{Reduction of Existing Algorithms into Lookahead Form}
In this section, we include the missing proofs for reducing the existing algorithms into its lookahead form.

\subsection{SNOO} \label{sec:compare_snoo}
SNOO \citep{kallusky2025snoo} (originated from DiLoCo \citep{douillard2023diloco} in federated learning) utilizes the $K$-step base optimizer update as a pseudo-gradient oracle for Nesterov's algorithm. In particular, for any base optimizer ${\cal A}_t(\cdot)$ and $K > 0$ inner steps, we can reduce the SNOO algorithm as the following lookahead algorithm with lookahead step size $\beta > 0$ by the derivation in Proof \ref{proof:snoo_reduction}.
\begin{equation}  \label{eq:snoo_eta_1}
    \begin{aligned}
    \textbf{for every}&~ t=nK, n = 0, ..., \lfloor T/K \rfloor,\\
        \tilde{\btheta}^{t+K} &= {\cal A}_{t+K-1}(~\cdots ~{\cal A}_{t+1}( {\cal A}_t(\bx^t - \beta {\bf b}^t) ) ) \\
        {\bf b}^{t+K} &=  \bx^{t} - \tilde{\btheta}^{t+K} \\
        \bx^{t+K} &= \tilde{\btheta}^{t+K} \\
    \textbf{end for;}~~&~~\textbf{return}~\bx^T - \beta {\bf b}^T.
    \end{aligned}
\end{equation}

\begin{prooft} \label{proof:snoo_reduction}
    The original SNOO algorithm can be summarized as follows:
    \begin{equation} 
    \begin{aligned}
    \textbf{for every}&~ t=nK, n = 0, ..., \lfloor T/K \rfloor,\\
        \tilde{\btheta}^{t+K} &= {\cal A}_{t+K-1}(~\cdots ~{\cal A}_{t+1}( {\cal A}_t(\btheta^t) ) ) \\
        {\bf b}^{t+K} &= \beta {\bf b}^t + (\btheta^{t} - \tilde{\btheta}^{t+K}) \\
        \btheta^{t+K} &= \btheta^t - \eta (\beta {\bf b}^{t+K} +  (\btheta^{t} - \tilde{\btheta}^{t+K})) \\
    \textbf{end for;}~~&~~\textbf{return}~\btheta^T.
    \end{aligned}
\end{equation}
with momentum $\beta >0$ and Nesterov's learning rate $\eta > 0$. The base optimizer ${\cal A}_t$ is applied for $K$ times consecutively from $\btheta^t$ to produce $\tilde{\btheta}^{t+K}$. However, it is non-trivial to understand the effect of Nesterov's momentum that is updated only after every $K$ stochastic gradient samples. To simplify the algorithm, we apply a change of variable $\btheta^t = \bx^t - \beta {\bf b}^t$ to revert the transformation of \cite{sutskever2013importance} into the following lookahead formulation:
\begin{equation} 
    \begin{aligned}
    \textbf{for every}&~ t=nK, n = 0, ..., \lfloor T/K \rfloor,\\
        \tilde{\btheta}^{t+K} &= {\cal A}_{t+K-1}(~\cdots ~{\cal A}_{t+1}( {\cal A}_t(\bx^t - \beta {\bf b}^t) ) ) \\
        {\bf b}^{t+K} &=  \bx^{t} - \tilde{\btheta}^{t+K} \\
        \bx^{t+K} &= \bx^t + (\beta - \eta \beta- \eta) {\bf b}^{t+K} - (\beta - \eta \beta ){\bf b}^t\\
    \textbf{end for;}~~&~~\textbf{return}~\bx^T - \beta {\bf b}^T.
    \end{aligned}
\end{equation}

Then, when $\eta = 1$, we have $\bx^{t+K} = \bx^t - {\bf b}^{t+K}$ and SNOO is reduced to \eqref{eq:snoo_eta_1}.

\hfill $\square$
\end{prooft}

\subsection{GPA} \label{sec:compare_gpa}
GPA \citep{defazio2025smoothing} adopts the primal-averaging framework which approximate Nesterov's lookahead. By the derivation in Proof \ref{proof:gpa_reduction}, we can reduce GPA into the following lookahead algorithm for learning rate $\gamma_p > 0$ and EMA rate $\beta > 0$: 

\begin{equation}  \label{eq:gpa_final}
    \begin{aligned}
    \textbf{for every}&~ t=0, ..., T-1,\\
        {\bf z}^{t+1} &= {\bf z}^t - \gamma_p \nabla f({\bf y}^t; \xi^t) \\
        {\bf x}^{t+1} &= \beta {\bf x}^t + ( 1 - \beta) {\bf z}^{t+1} \\
        {\bf y}^{t+1} &= {\bf y}^{t} + (1- \beta)({\bf z}^{t+1} - {\bf z}^t) + \beta(1-\beta)({\bf z}^{t+1} - {\bf x}^t)\\
    \textbf{end for;}~~&~~\textbf{return}~\bx^T.
    \end{aligned}
\end{equation}

\begin{prooft} \label{proof:gpa_reduction}
We summarize the GPA algorithm as follows:
\begin{equation} 
    \begin{aligned}
    \textbf{for every}&~ t=0, ..., T-1,\\
        {\bf y}^{t} &= \beta_y {\bf x}^t + ( 1 -\beta_y) {\bf z}^t \\
        {\bf z}^{t+1} &= {\bf z}^t + {\cal A}_t({\bf y}^t) - {\bf y}^t \\
        {\bf x}^{t+1} &= \beta_x {\bf x}^t + ( 1 - \beta_x) {\bf z}^{t+1} \\
    \textbf{end for;}~~&~~\textbf{return}~\bx^T.
    \end{aligned}
\end{equation}
for momentum $\beta_x \in  [0, 1), \beta_y \in [0, 1]$. 
Now apply \cite[Proposition 1]{defazio2025smoothing} to obtain the following equivalent algorithm when ${\cal A}_t({\bf y}) = {\bf y} - \gamma_p \nabla f({\bf y}; \xi^t)$:
\begin{equation} 
    \begin{aligned}
    \textbf{for every}&~ t=0, ..., T-1,\\
        {\bf z}^{t+1} &= {\bf z}^t - \gamma_p \nabla f({\bf y}^t; \xi^t) \\
        {\bf x}^{t+1} &= \beta {\bf x}^t + ( 1 - \beta) {\bf z}^{t+1} \\
        {\bf b}^{t} &= \frac{1}{\gamma_m} ({\bf x}^t - {\bf x}^{t+1}) = \beta {\bf b}^{t-1} + \nabla f({\bf y}^t; \xi^t) \\
        {\bf y}^{t+1} &= {\bf y}^{t} - \gamma_m (\beta {\bf b}^t + \nabla f({\bf y}^t; \xi^t))\\
    \textbf{end for;}~~&~~\textbf{return}~\bx^T.
    \end{aligned}
\end{equation}
with $\beta = \beta_x = \beta_y$ and $\gamma_m = (1 - \beta)\gamma_p$. To obtain \eqref{eq:gpa_final}, we first prove that ${\bf x}^t = {\bf z}^t +\gamma_p  \beta {\bf b}^{t-1}$ because
\begin{align}
    {\bf z}^t +\gamma_p  \beta {\bf b}^{t-1} &= {\bf z}^t + \frac{\beta}{1 - \beta} ({\bf x}^{t-1} - {\bf x}^t)\\
    &={\bf z}^t + \frac{\beta}{1 - \beta} ({\bf x}^{t-1} - \beta {\bf x}^{t-1} - (1 - \beta){\bf z}^t) \\
     &=\beta {\bf x}^{t-1} + (1 - \beta) {\bf z}^t = {\bf x}^t.
\end{align}
Then, to simplify ${\bf y}^{t+1}$, we observe that
\begin{align}
    {\bf b}^t &= \beta {\bf b}^{t-1} + \nabla f({\bf y}^t; \xi^t) =\beta {\bf b}^{t-1} + \frac{1}{\gamma_p} ({\bf z}^t - {\bf z}^{t+1}) = \frac{1}{\gamma_p}({\bf x}^t - {\bf z}^{t+1}),
\end{align}
therefore,
\begin{align}
    {\bf y}^{t+1} &= {\bf y}^t - \gamma_m\beta{\bf b}^t - \gamma_m \nabla f({\bf y}^t; \xi^t) \\
    &={\bf y}^t - \frac{\gamma_m \beta}{\gamma_p}({\bf x}^t - {\bf z}^{t+1}) - \frac{\gamma_m}{\gamma_p} ({\bf z}^t - {\bf z}^{t+1}) \\
    &= {\bf y}^{t} + (1- \beta)({\bf z}^{t+1} - {\bf z}^t) + \beta(1-\beta)({\bf z}^{t+1} - {\bf x}^t).
\end{align}
\hfill $\square$
\end{prooft}

\section{Convergence Proof}
{\bf Notation.} We denote $\| \cdot \| = \| \cdot \|_2$ as the vector 2-norm.
\subsection{Strongly Convex} \label{app:proof_strong_cvx}
Under an appropriate choice of Lyapunov function ${\cal E}_t$ where
\begin{equation} \label{eq:lyap_strong_cvx}
    {\cal E}_t := f(\bx^t) - f(\bx^\star) + \frac{\mu}{2} \| {\bf v}^t - \bx^\star \|^2
\end{equation}
where ${\bf v}^t := \bx^t + c{\bf m}^t$ for some $c > 0$, we provide a proof of linear convergence towards the optimal solution as follows.

First, we expand the one step evolution of every variable as follows:
\begin{align}
    \bx^{t+1} = \by^t - \alpha \nabla f(\by^t) = \bx^t + \beta\bm^t - \alpha \nabla f(\by^t) ,\label{eq:x_one_step}
\end{align}
\begin{align}
    \bm^{t+1} &= \gamma \bm^t + (1-\gamma)(\bx^{t+1} - \bx^t) \\
    &\stackrel{\eqref{eq:x_one_step}}{=}\gamma \bm^t + (1-\gamma)(\beta \bm^t - \alpha \nabla f(\by^t)) \\
    &= (\gamma + (1-\gamma)\beta) \bm^t - (1-\gamma)\alpha \nabla f(\by^t). \label{eq:m_one_step}
\end{align}
Then, the one step evolution of $\bv^t$ can be derived as
\begin{align}
    \bv^{t+1} &= \bx^{t+1} + c\bm^{t+1} \\
    &= \bx^t + \beta\bm^t - \alpha \nabla f(\by^t) + c \left[ (\gamma + (1-\gamma)\beta) \bm^t - (1-\gamma)\alpha \nabla f(\by^t) \right] \\
    &= \bx^t + [\beta + c(\gamma + (1-\gamma)\beta)] \bm^t - \alpha [1 + c(1-\gamma)]\nabla f(\by^t). \label{eq:v_lhs}
\end{align}
Now we wish to find some $c > 0$ such that the above equation can be equivalently reduced to
\begin{equation}
    \bv^{t+1} = (1 - r) \bv^t + r\left(\by^t - \frac{1}{\mu} \nabla f(\by^t) \right) \label{eq:v_hypo}
\end{equation}
for some $r > 0$. In this regard, we expand the R.H.S. of \eqref{eq:v_hypo} and obtain
\begin{align}
&(1 - r) (\bx^t + c \bm^t) + r\left(\bx^t + \beta \bm^t - \frac{1}{\mu} \nabla f(\by^t) \right) \\
&=\bx^t + [(1-r)c + r\beta] \bm^t - \frac{r}{\mu}\nabla f(\by^t). \label{eq:v_hypo_rhs}
\end{align}
Then we are left with matching the coefficients between \eqref{eq:v_lhs} and \eqref{eq:v_hypo_rhs}, i.e., we find a set of $c, r, \beta$ such that
\begin{align}
    &\beta + c(\gamma + (1-\gamma)\beta) = (1-r) c + r\beta \label{eq:m_coe}
\end{align}
and 
\begin{align}
    \alpha [1 + c(1-\gamma)] = \frac{r}{\mu} \Leftrightarrow r = \alpha \mu [1 + c(1-\gamma)] .\label{eq:grad_coe}
\end{align}
By choosing
\begin{equation}
    c = \frac{1-\gamma - r}{(1-\gamma)r}, \label{eq:c_choice}
\end{equation}
we deduce that
\begin{align}
    r &= \alpha \mu\left( 1+ \frac{1-\gamma -r}{r}\right) =\alpha \mu \frac{1-\gamma}{r} \\
    &\Leftrightarrow r = \sqrt{\alpha \mu (1-\gamma)} \label{eq:r_choice}
\end{align}
will match the gradient term coefficient as written in \eqref{eq:grad_coe}. Meanwhile, back to the coefficient of $\bm^t$ term in \eqref{eq:m_coe}, we get
\begin{align}
    &\beta + \frac{1-\gamma-r}{(1-\gamma) r} (\gamma + (1-\gamma) \beta)) = (1 - r) \frac{1-\gamma - r}{(1-\gamma)r} + r \beta \\
    &\Leftrightarrow \beta (1+ \frac{1-\gamma - r}{r} - r) = \frac{1-r}{r} \cdot \frac{1-\gamma - r}{1-\gamma} - \gamma \cdot \frac{1-\gamma -r}{(1-\gamma)r} \\
    &\Leftrightarrow \beta (\frac{1-\gamma }{r} - r) = \frac{(1-\gamma-r)^2}{r(1-\gamma)} \\
    &\Leftrightarrow \beta (1-\gamma - r^2) = \frac{(1-\gamma-r)^2}{(1-\gamma)} \\
    &\Leftrightarrow \beta  = \frac{(1-\gamma-r)^2}{(1-\gamma)(1-\gamma - r^2)} \label{eq:beta_choice}
\end{align}
will match the $\bm^t$ term coefficient as well. Next, using \eqref{eq:v_hypo} and the equality $\| (1-\theta) {\bf u} + \theta {\bf w} \|^2 = (1-\theta) \| {\bf u} \|^2 + \theta \| {\bf w} \|^2 - \theta (1-\theta) \| {\bf u} - {\bf w} \|^2$ for any ${\bf u, w} \in \mathbb{R}^d$, $\theta \in [0,1]$, we observe the evolution of the Lyapunov function by expanding
\begin{align}
    &\| \bv^{t+1} - \bx^\star \|^2 \\
    &= \| (1-r) (\bv^t-\bx^\star) + r (\by^t - \bx^\star - \frac{1}{\mu} \nabla f(\by^t)) \|^2 \\
    &=(1-r) \| \bv^t-\bx^\star \|^2 + r \| \by^t - \bx^\star - \frac{1}{\mu} \nabla f(\by^t)\|^2 - r(1-r) \| \bv^t - \by^t + \frac{1}{\mu} \nabla f(\by^t) \|^2. \label{eq:vx_1}
\end{align}
Notice that
\begin{align}
    \| \by^t - \bx^\star - \frac{1}{\mu} \nabla f(\by^t)\|^2 = \| \by^t - \bx^\star\|^2 + \frac{1}{\mu^2} \|\nabla f(\by^t) \|^2 -\frac{2}{\mu} \dotp{\nabla f(\by^t)}{\by^t - \bx^\star}
\end{align}
Under $\mu$-strong convexity, we can combine the above with
\begin{align}
    &f(\bx^\star) \ge  f(\by^t) + \dotp{\nabla f(\by^t)}{\bx^\star - \by^t}  + \frac{\mu}{2} \|\by^t - \bx^\star \|^2 \\
    &\Leftrightarrow - \dotp{\nabla f(\by^t)}{\by^t - \bx^\star} \leq f(\bx^\star) - f(\by^t) - \frac{\mu}{2} \| \by^t - \bx^\star \|^2
\end{align}
and obtain
\begin{align}
    \| \by^t - \bx^\star - \frac{1}{\mu} \nabla f(\by^t)\|^2 \leq \frac{1}{\mu^2} \|\nabla f(\by^t) \|^2 -\frac{2}{\mu}(f(\by^t)  - f(\bx^\star) ). \label{eq:yx_star_grad}
\end{align}
Putting together \eqref{eq:vx_1} and \eqref{eq:yx_star_grad} yields
\begin{align}
    \| \bv^{t+1} - \bx^\star \|^2  &\leq (1-r) \| \bv^t-\bx^\star \|^2 + \frac{r}{\mu^2} \|\nabla f(\by^t) \|^2 -\frac{2r}{\mu}(f(\by^t)  - f(\bx^\star) )\\
    &\quad - r(1-r) \| \bv^t - \by^t + \frac{1}{\mu} \nabla f(\by^t) \|^2. \label{eq:lya_1}
\end{align}
Meanwhile, under $L$-smooth assumption, 
\begin{align}
    f(\bx^{t+1}) - f(\bx^\star) &\leq f(\by^t) + \dotp{\nabla f(\by^t)}{\bx^{t+1} - \by^t} + \frac{L}{2} \| \bx^{t+1} -\by^t \|^2 -f(\bx^\star) \\
    &= f(\by^t) - \alpha \| \nabla f(\by^t) \|^2 + \frac{\alpha^2 L}{2} \| \nabla f(\by^t) \|^2 - f(\bx^\star) \\
    &\leq f(\by^t) -f(\bx^\star) - \frac{\alpha}{2} \| \nabla f(\by^t) \|^2 \label{eq:lya_2}
\end{align}
when $\alpha^2 L / 2 \leq \alpha / 2 \Leftrightarrow \alpha \leq 1/L$. Combining \eqref{eq:lya_1} and \eqref{eq:lya_2} together gives us the recursion of the Lyapunov function as
\begin{align}
    {\cal E}_{t+1} &\leq (1-r)(f(\by^t) -f(\bx^\star)) + (\frac{r}{2\mu} - \frac{\alpha}{2} )\| \nabla f(\by^t) \|^2 + \frac{\mu(1-r)}{2} \| \bv^t-\bx^\star \|^2 \\
    &\quad - \frac{\mu r(1-r)}{2} \| \bv^t - \by^t + \frac{1}{\mu} \nabla f(\by^t) \|^2 \\
    &= (1-r){\cal E}_t + (1-r)(f(\by^t) - f(\bx^t)) + (\frac{r}{2\mu} - \frac{\alpha}{2} )\| \nabla f(\by^t) \|^2\\
    &\quad - \frac{\mu r(1-r)}{2} \| \bv^t - \by^t + \frac{1}{\mu} \nabla f(\by^t) \|^2 \\
    &\leq (1-r){\cal E}_t + (1-r)\dotp{\nabla f(\by^t)}{\by^t - \bx^t} - \frac{\mu (1-r)}{2} \| \by^t - \bx^t\|^2\\
    &\quad  + (\frac{r}{2\mu} - \frac{\alpha}{2} )\| \nabla f(\by^t) \|^2  - \frac{\mu r(1-r)}{2} \| \bv^t - \by^t + \frac{1}{\mu} \nabla f(\by^t) \|^2 ,
\end{align}
where the last inequality is due to $\mu$-strong convexity.
Now recall by the definition of $\bv^t = \bx^t + c\bm^t$, we have
\begin{align}
    \by^t &= \bx^t + \beta \bm^t = \left(1 - \beta/c\right) \bx^t + \beta/c \cdot \bv^t \\
    \Leftrightarrow \bv^t &= c/\beta \cdot \by^t - \left(c/\beta -1\right)\bx^t \\
    \Leftrightarrow \bv^t -\by^t &= \left(c/\beta - 1\right) (\by^t -\bx^t).    
\end{align}
This allows us to expand
\begin{align}
    \| \bv^t - \by^t + \frac{1}{\mu} \nabla f(\by^t) \|^2 &= \left(c/\beta - 1\right)^2 \| \by^t - \bx^t \|^2 + \frac{1}{\mu^2} \| \nabla f(\by^t)\|^2  \\
    &\quad + \frac{2}{\mu}\left(c/\beta - 1\right)\dotp{\by^t - \bx^t}{\nabla f(\by^t)}.
\end{align}
Putting it back to the recursion of the Lyapunov function, we get
\begin{align}
    {\cal E}_{t+1} &\leq (1-r){\cal E}_t + [(1-r) - (c/\beta - 1) r (1-r)]\dotp{\nabla f(\by^t)}{\by^t - \bx^t} \\
    &\quad -\left( \frac{\mu (1-r)}{2} + (c/\beta - 1)^2 \frac{\mu r(1-r)}{2}\right) \| \by^t - \bx^t\|^2\\
    &\quad  -\left(-\frac{r}{2\mu} + \frac{\alpha}{2} + \frac{r(1-r)}{2\mu} \right)\| \nabla f(\by^t) \|^2.
\end{align}
By $\dotp{\nabla f(\by^t)}{\by^t - \bx^t} \leq \|\nabla f(\by^t)\| \| \by^t - \bx^t\|$, we rewrite the residual by defining $C := (1-r)(1 - r(c/\beta - 1))$, $A:=\frac{\mu(1-r)}{2}(1 + r(c/\beta-1)^2)$, $B:=\left(-\frac{r}{2\mu} + \frac{\alpha}{2} + \frac{r(1-r)}{2\mu} \right)$ so that
\begin{align}
    {\cal E}_{t+1} \leq (1-r) {\cal E}_t + |C| \| \nabla f(\by^t) \| \cdot \| \by^t - \bx^t \| - A \| \by^t - \bx^t \|^2 - B \| \nabla f(\by^t) \|^2.
\end{align}
To achieve ${\cal E}_{t+1} \leq (1-r){\cal E}_t$, our last step is to show 
\begin{align}
     &|C| \| \nabla f(\by^t) \| \cdot \| \by^t - \bx^t \| - A \| \by^t - \bx^t \|^2 - B \| \nabla f(\by^t) \|^2 \leq 0 \\
     &\Leftrightarrow A\left( \| \by^t - \bx^t \|^2 - \frac{|C|}{A}  \cdot \| \nabla f(\by^t) \| \cdot \| \by^t - \bx^t \| + (\frac{|C|}{2A} \| \nabla f(\by^t) \|)^2 \right) \\
     &\quad + (- \frac{C^2}{4A}  + B ) \| \nabla f(\by^t) \|^2 \ge 0 \\
     &\Leftrightarrow A\left( \| \by^t - \bx^t \| - \frac{|C|}{2A}\| \nabla f(\by^t) \| \right)^2 + (B- \frac{C^2}{4A})  \| \nabla f(\by^t) \|^2 \ge 0.
\end{align}
Since it is obvious that $A\ge 0$, we are left with verifying $B-C^2/(4A) \ge 0 \Leftrightarrow 4AB-C^2 \ge 0$. We begin by simplifying each constant as
\begin{equation}
    B = \frac{\alpha}{2} - \frac{r^2}{2\mu} \stackrel{\eqref{eq:r_choice}}{=}\frac{\alpha}{2} - \frac{\alpha \mu(1-\gamma)}{2\mu} = \frac{\alpha\gamma}{2} \stackrel{\eqref{eq:r_choice}}{=} \frac{r^2 \gamma}{2(1-\gamma)\mu},
\end{equation}
\begin{align}
    &c/\beta \stackrel{\eqref{eq:c_choice},\eqref{eq:beta_choice}}{=}\frac{1-\gamma-r}{(1-\gamma)r} \cdot \frac{(1-\gamma)(1-\gamma-r^2)}{(1-\gamma-r)^2} = \frac{1-\gamma-r^2}{r(1-\gamma-r)} \\
    &\Rightarrow c/\beta - 1 = \frac{(1-r)(1-\gamma)}{r(1-\gamma-r)} \label{eq:c_beta_1}
\end{align}
\begin{align}
    C &\stackrel{\eqref{eq:c_beta_1}}{=}(1-r)(1 -\frac{(1-r)(1-\gamma)}{1-\gamma-r}) \\
    &= \frac{1-r}{1-\gamma - r} \cdot (1-\gamma -r -1 + r + \gamma - r\gamma)\\
    &= - \frac{\gamma r(1-r)}{1-\gamma - r} ,
\end{align}
\begin{align}
A &\stackrel{\eqref{eq:c_beta_1}}{=} \frac{\mu(1-r)}{2} (1 + r\cdot \frac{(1-r)^2(1-\gamma)^2}{r^2 (1-\gamma-r)^2}) \\
&= \frac{\mu(1-r)}{2r (1-\gamma-r)^2} (r (1-\gamma-r)^2 + (1-r)^2(1-\gamma)^2).
\end{align}
Therefore,
\begin{align}
    4AB-C^2 &= \frac{r \gamma (1-r)}{(1-\gamma) (1-\gamma-r)^2} (r (1-\gamma-r)^2 + (1-r)^2(1-\gamma)^2) - \frac{\gamma^2 r^2 (1-r)^2}{(1-\gamma-r)^2}  \\
    &=\frac{r\gamma(1-r)}{(1-\gamma-r)^2}\left[ \frac{r (1-\gamma-r)^2}{1-\gamma} + (1-r)^2(1-\gamma) - \gamma r (1-r) \right].
\end{align}
Using
\begin{align}
    &r (1-\gamma-r)^2 + (1-r)^2(1-\gamma)^2 - (1-\gamma)\gamma r (1-r) \\
    &= r (1-\gamma-r)^2 + (1-r) (1-\gamma)[(1-r)(1-\gamma) - \gamma r] \\
    &= r(1-\gamma -r)^2 + (1-r)(1-\gamma)(1-r-\gamma) \\
    &= (1-\gamma-r)(1-\gamma - r^2),
\end{align}to c
we obtain
\begin{align}
    4AB-C^2 = \frac{r\gamma(1-r)(1-\gamma-r^2)}{(1-\gamma-r)(1-\gamma)}.
\end{align}
    Since $0\leq r \leq 1, 0 \leq \gamma \leq 1$, by plugging in $r = \sqrt{\alpha \mu (1-\gamma) }$ we have $1-\gamma -r^2 \ge 0  \Leftrightarrow \gamma \leq 1$. Meanwhile, $1-\gamma - r \ge0 \Leftrightarrow \gamma \leq 1- \alpha \mu$. Therefore, we have $4AB - C^2 \ge 0$ and this completes the proof.
\hfill $\square$

\subsection{General Convex} \label{app:proof_cvx}
For general convex objective function, we aim to adopt another choice of Lyapunov function $\Phi_t$ where
\begin{equation} \label{eq:lyap_cvx}
    \Phi_t:= A_t(f(\bx^t) - f(\bx^\star)) + \frac{1}{2} \|\bx^t + c_t {\bf m}^t - \bx^\star \|^2_2
\end{equation}
with some $A_t, c_t > 0$ determined later. Then, we can guarantee a sublinear convergence. Below we present the proof.

Define the auxiliary sequence
$
\mathbf{v}^{t}:=\mathbf{x}^{t}+c_t\mathbf{m}^{t}.
$
We first expand the one step evolution of every variable as follows:
\begin{align}
\mathbf{x}^{t+1}
&=
\mathbf{x}^{t}+\beta_t \mathbf{m}^{t}-\frac1L \nabla f(\mathbf{y}^{t}),\label{eq:x_one_step_convex}\\
\mathbf{m}^{t+1} 
&\stackrel{\eqref{eq:x_one_step_convex}}{=}
\gamma \mathbf{m}^{t}
+
(1-\gamma)(\mathbf{x}^{t+1}-\mathbf{x}^{t}) \\
&=
\bigl(\gamma+(1-\gamma)\beta_t\bigr)\mathbf{m}^{t}
-
\frac{1-\gamma}{L}\nabla f(\mathbf{y}^{t}).
\end{align}
Then, the one step evolution of $\bv^t$ can be derived as
\begin{align}
\mathbf{v}^{t+1}
&=\mathbf{x}^{t+1}+c_{t+1}\mathbf{m}^{t+1} \\
&=
\mathbf{x}^{t}+\beta_t \mathbf{m}^{t}-\frac1L \nabla f(\mathbf{y}^{t})
+
c_{t+1}
\left[
\bigl(\gamma+(1-\gamma)\beta_t\bigr)\mathbf{m}^{t}
-
\frac{1-\gamma}{L}\nabla f(\mathbf{y}^{t})
\right] \\
&=
\mathbf{x}^{t}
+
\left[
\beta_t+c_{t+1}\bigl(\gamma+(1-\gamma)\beta_t\bigr)
\right]\mathbf{m}^{t}
-
\frac{1+(1-\gamma)c_{t+1}}{L}\nabla f(\mathbf{y}^{t}).
\end{align}
By the definition of \(\beta_t\),
\begin{align}
\beta_t\bigl(1+(1-\gamma)c_{t+1}\bigr)+\gamma c_{t+1}=c_t.
\end{align}
Thus
\begin{align}
\mathbf{v}^{t+1}
&=
\mathbf{x}^{t}+c_t\mathbf{m}^{t}
-
\frac{1+(1-\gamma)c_{t+1}}{L}\nabla f(\mathbf{y}^{t}) \\
&=
\mathbf{v}^{t}-\eta_t \nabla f(\mathbf{y}^{t}), \label{eq:vt_dyn_convex}
\end{align}
where
$
\eta_t:=(1+(1-\gamma)c_{t+1})/L.
$

Since \(\mathbf{v}^{t}-\mathbf{x}^{t}=c_t\mathbf{m}^{t}\), we have
\begin{align}
\mathbf{y}^{t}
&=
\mathbf{x}^{t}+\beta_t\mathbf{m}^{t} 
=
\mathbf{x}^{t}+\frac{\beta_t}{c_t}(\mathbf{v}^{t}-\mathbf{x}^{t}) 
=
\left(1-\frac{\beta_t}{c_t}\right)\mathbf{x}^{t}
+
\frac{\beta_t}{c_t}\mathbf{v}^{t}.
\end{align}

We now find a sequence $A_t$ so that the following holds:
\begin{align}\label{eq:cond_at_convex}
A_{t+1}\left(1-\frac{\beta_t}{c_t}\right)\le A_t,
\qquad
A_{t+1}\frac{\beta_t}{c_t}=\eta_t.
\end{align}

In fact, define a sequence \(A_t=a_t/L\) as follows:
\begin{align}\label{eq:at_def_convex}
a_{t+1}:=\frac{c_tq_t^2}{c_t-\gamma c_{t+1}},
\qquad t\ge 0,
\end{align}
where \begin{align}\label{eq:qt_def_convex}
q_t:=1+(1-\gamma)c_{t+1}.
\end{align}
Specifically, for $t=0$, choose \(a_0\ge a_1-q_0\) (for example, choose $a_0=a_1$). For the given choice of \(c_t\), one can easily verify
\begin{align}
c_t-\gamma c_{t+1}
=
(1-\gamma)\left(1+\frac{t}{4}\right),
\qquad
q_t
=
1+(1-\gamma)c_{t+1}
=
\frac54+(1-\gamma)\left(1+\frac{t}{4}\right).
\end{align}
Therefore, by \eqref{eq:at_def_convex}, \eqref{eq:qt_def_convex} as well as the definition of $\beta_t$ and $\eta_t$,
\begin{align}
A_{t+1}\frac{\beta_t}{c_t}
=
\frac{a_{t+1}}{L}\cdot \frac{\beta_t}{c_t} 
=
\frac{1}{L}
\frac{c_tq_t^2}{c_t-\gamma c_{t+1}}
\cdot
\frac{c_t-\gamma c_{t+1}}{c_t q_t}
=
\frac{q_t}{L} 
=
\eta_t.
\end{align}
We now show $A_{t+1}\left(1-\frac{\beta_t}{c_t}\right)\le A_t$, or equivalently, $a_{t+1}-q_t\le a_t$. When $t=0$, the inequality holds due to our choice of $a_0$. For \(t\ge 1\), put $s_t:=1+\frac{t}{4}$. Then a direct calculation gives
\begin{align}
a_t-(a_{t+1}-q_t)
=
\frac{
8(1-\gamma)^3 s_t(16s_t^2-1)
+
36(1-\gamma)^2 s_t(4s_t-1)
+
25\gamma
}{
64(1-\gamma)^2 s_t(4s_t-1)
}.
\end{align}
Since \(t\ge 1\), we have \(s_t\ge 5/4\). Also \(0<1-\gamma\le 1\).
Therefore every term in the numerator is nonnegative and the denominator is
positive. Hence $a_t-(a_{t+1}-q_t)\ge 0$. Thus \eqref{eq:cond_at_convex} holds.

Given the choice of $A_t$, we now prove the Lyapunov function $\Phi_t$ decreases.

Since \(f\) is \(L\)-smooth and
$
\mathbf{x}^{t+1}
=
\mathbf{y}^{t}-\frac1L \nabla f(\mathbf{y}^{t}),
$
we have
\begin{align}\label{eq_lsmooth_convex}
f(\mathbf{x}^{t+1})
\le
f(\mathbf{y}^{t})-\frac{1}{2L}\|\nabla f(\mathbf{y}^{t})\|^2.
\end{align}
Furthermore, since
\begin{align}
\mathbf{y}^{t}
=
\left(1-\frac{\beta_t}{c_t}\right)\mathbf{x}^{t}
+
\frac{\beta_t}{c_t}\mathbf{v}^{t},
\end{align}
we have by convexity that
\begin{align}\label{eq:three_point_convex}
f(\mathbf{y}^{t})-f(\mathbf{x}^\star)
&=
\left(1-\frac{\beta_t}{c_t}\right)
\bigl(f(\mathbf{y}^{t})-f(\mathbf{x}^\star)\bigr)
+
\frac{\beta_t}{c_t}
\bigl(f(\mathbf{y}^{t})-f(\mathbf{x}^\star)\bigr)
\nonumber\\
&\le
\left(1-\frac{\beta_t}{c_t}\right)
\left[
f(\mathbf{x}^{t})-f(\mathbf{x}^\star)
-
\left\langle
\nabla f(\mathbf{y}^{t})\middle|
\mathbf{x}^{t}-\mathbf{y}^{t}
\right\rangle
\right]
+
\frac{\beta_t}{c_t}
\left\langle
\nabla f(\mathbf{y}^{t})\middle|
\mathbf{y}^{t}-\mathbf{x}^\star
\right\rangle
\nonumber\\
&=
\left(1-\frac{\beta_t}{c_t}\right)
\bigl(f(\mathbf{x}^{t})-f(\mathbf{x}^\star)\bigr)
+
\left\langle
\nabla f(\mathbf{y}^{t})\middle|
\frac{\beta_t}{c_t}(\mathbf{y}^{t}-\mathbf{x}^\star)
-
\left(1-\frac{\beta_t}{c_t}\right)(\mathbf{x}^{t}-\mathbf{y}^{t})
\right\rangle
\nonumber\\
&=
\left(1-\frac{\beta_t}{c_t}\right)
\bigl(f(\mathbf{x}^{t})-f(\mathbf{x}^\star)\bigr)
+
\frac{\beta_t}{c_t}
\left\langle
\nabla f(\mathbf{y}^{t})\middle|
\mathbf{v}^{t}-\mathbf{x}^\star
\right\rangle .
\end{align}
Multiplying \eqref{eq:three_point_convex} by \(A_{t+1}\) and then using \eqref{eq:cond_at_convex} and \eqref{eq_lsmooth_convex},
we obtain
\begin{align}\label{eq:f_contraction_convex}
A_{t+1}\bigl(f(\mathbf{x}^{t+1})-f(\mathbf{x}^\star)\bigr)
&\le
A_t\bigl(f(\mathbf{x}^{t})-f(\mathbf{x}^\star)\bigr)
+
\eta_t\langle \nabla f(\mathbf{y}^{t}),\mathbf{v}^{t}-\mathbf{x}^\star\rangle 
-
\frac{A_{t+1}}{2L}\|\nabla f(\mathbf{y}^{t})\|^2.
\end{align}
On the other hand, by \eqref{eq:vt_dyn_convex} we have
\begin{align}\label{eq:v_dist_convex}
\frac12\|\mathbf{v}^{t+1}-\mathbf{x}^\star\|^2
=
\frac12\|\mathbf{v}^{t}-\mathbf{x}^\star\|^2
-
\eta_t\left\langle \nabla f(\mathbf{y}^{t})\middle|
\mathbf{v}^{t}-\mathbf{x}^\star\right\rangle
+
\frac{\eta_t^2}{2}\|\nabla f(\mathbf{y}^{t})\|^2.
\end{align}
Adding \eqref{eq:f_contraction_convex} and \eqref{eq:v_dist_convex} yields
\begin{align}\label{eq:phi_contraction_convex}
\Phi_{t+1}
\le
\Phi_t
-
\frac12
\left(
\frac{A_{t+1}}{L}-\eta_t^2
\right)
\|\nabla f(\mathbf{y}^{t})\|^2.
\end{align}
It remains to check that
\begin{align}
\frac{A_{t+1}}{L}\ge \eta_t^2.
\end{align}
Since \(A_{t+1}=a_{t+1}/L\) and \(\eta_t=q_t/L\), this is equivalent to
$a_{t+1}\ge q_t^2$, which apparently holds by our choice of $a_{t+1}$ in \eqref{eq:at_def_convex}. Combining with \eqref{eq:phi_contraction_convex}, we show that
\begin{align}
\Phi_{t+1}\le \Phi_t.
\end{align}
Consequently,
$
A_t\bigl(f(\mathbf{x}^{t})-f(\mathbf{x}^\star)\bigr)
\le
\Phi_t
\le
\Phi_0
$, and therefore
\begin{align}\label{eq:f_bound_convex}
f(\mathbf{x}^{t})-f(\mathbf{x}^\star)
\le
\frac{\Phi_0}{A_t}
=
\frac{L\Phi_0}{a_t}.
\end{align}

Next, we lower bound \(a_t\). For \(t\ge 1\), by \eqref{eq:at_def_convex} we have
\begin{align}
a_t
=
\frac{c_{t-1}q_{t-1}^2}{c_{t-1}-\gamma c_t}.
\end{align}
Recalling $
s_{t-1}=1+\frac{t-1}{4},
$
a direct calculation can verify
\begin{align}
c_{t-1}\ge s_{t-1},
\qquad
q_{t-1}\ge (1-\gamma)s_{t-1},
\qquad
c_{t-1}-\gamma c_t=(1-\gamma)s_{t-1}.
\end{align}
Thus
\begin{align}\label{eq:at_bound_convex}
a_t
\ge
\frac{s_{t-1}\bigl((1-\gamma)s_{t-1}\bigr)^2}{(1-\gamma)s_{t-1}} 
=
(1-\gamma)s_{t-1}^2\ge \frac{(1-\gamma)t^2}{16}.
\end{align}
Therefore by \eqref{eq:f_bound_convex} and \eqref{eq:at_bound_convex}, we obtain
\begin{align}\label{eq:t2-dependency}
f(\mathbf{x}^{t})-f(\mathbf{x}^\star)
\le
\frac{16L\Phi_0}{(1-\gamma)t^2}.
\end{align}
\eqref{eq:t2-dependency} provides the $O(1/t^2)$ rate of the algorithm. We further upper bound $\Phi_0$ to provide the $\gamma$-dependency. Note that $\mathbf m^0=0$. Since $f$ is convex and $L$-smooth and
$\mathbf x^\star$ is a minimizer, we have
\begin{equation}\label{eq:f-L-smooth_xstar}
    f(\mathbf x^0)-f(\mathbf x^\star)
\le
\frac{L}{2}\|\mathbf x^0-\mathbf x^\star\|_2^2.
\end{equation}
Combining \eqref{eq:lyap_cvx} and \eqref{eq:f-L-smooth_xstar} yields
\begin{equation}\label{eq:phi_l2_bound}
    \Phi_0
=
\frac{a_0}{L}\bigl(f(\mathbf x^0)-f(\mathbf x^\star)\bigr)
+
\frac12\|\mathbf x^0-\mathbf x^\star\|_2^2
\le
\frac{a_0+1}{2}
\|\mathbf x^0-\mathbf x^\star\|_2^2.
\end{equation}
Now we calculate $a_0$. By the definition of $c_t$, $a_t$ and $q_t$ (recall \eqref{eq:at_def_convex} and \eqref{eq:qt_def_convex}), we have
\begin{equation}
    c_0=1+\frac{\gamma}{4(1-\gamma)}, \qquad c_1=\frac{5}{4}+\frac{\gamma}{4(1-\gamma)}, \qquad q_0=1+(1-\gamma)c_1, \qquad a_1=\frac{c_0q_0^2}{c_0-\gamma c_1}.
\end{equation}
We need $a_0\geq a_1-q_0$. By simply choosing $a_0=a_1$, a direct calculation yields
\begin{equation}\label{eq:a0_bound}
    a_0=a_1=\frac{(4-3\gamma)(9-4\gamma)^2}{64(1-\gamma)^2}\leq \frac{81}{16(1-\gamma)^2}.
\end{equation}
Therefore, by combining \eqref{eq:t2-dependency}, \eqref{eq:phi_l2_bound} and \eqref{eq:a0_bound}, we get
\begin{equation}
\begin{aligned}
    f(\mathbf{x}^{t})-f(\mathbf{x}^\star)
\leq&
\frac{16L}{(1-\gamma)t^2}\cdot \frac{a_0+1}{2}
\|\mathbf x^0-\mathbf x^\star\|_2^2\\
\leq& \frac{16L}{(1-\gamma)t^2}\cdot \frac{\|\mathbf x^0-\mathbf x^\star\|_2^2}{2}\cdot \frac{97}{16(1-\gamma)^2}\\
=&\frac{97L}{2(1-\gamma)^3t^2}\|\mathbf x^0-\mathbf x^\star\|_2^2.
\end{aligned}
\end{equation}
This completes the proof.
\hfill $\square$

\newpage
\section{Additional Experiments Details}
\subsection{Validation Loss at Lookahead Positions}
\begin{figure}[htb!]
    \centering
    \includegraphics[width=0.4\linewidth]{figures/nanogpt_lookahead_legend.pdf} \\
    \includegraphics[width=0.325\linewidth]{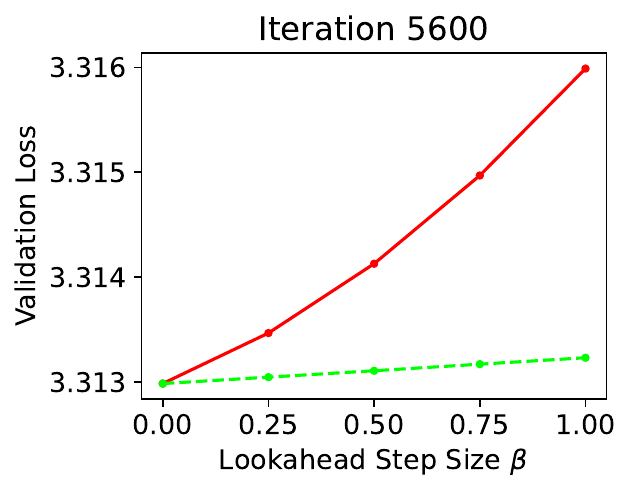}
    \includegraphics[width=0.325\linewidth]{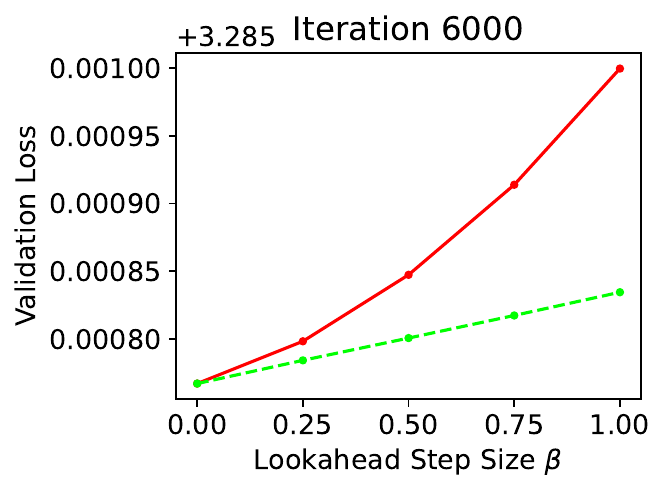}
    \includegraphics[width=0.325\linewidth]{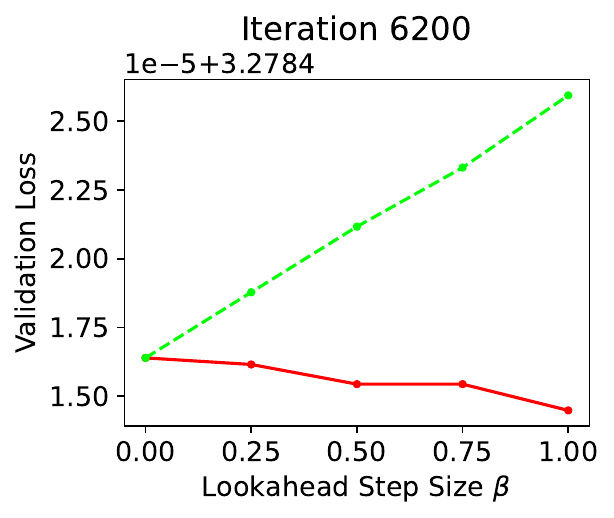}
    \caption{Under the NanoGPT setup, we measure the validation loss in lookahead positions $\by^t = \bx^t + \beta {\bf d}^t$ at iterations $t=5600, 6000, 6200$ using ({\color{red}{Solid Red}}) $\by^t = \bx^t + \beta \cdot (\bx^t - \bx^{t-1})$ and ({\color{green!50!black}Dashed Green}) $\by^t = \bx^t + \beta \cdot {\bf m}^t$ where ${\bf m}^t = \gamma {\bf m}^{t-1} + (1-\gamma) (\bx^t - \bx^{t-1})$ with $\gamma=0.995$.
    }
    \label{fig:nanogpt_lookahead_convergence}
\end{figure}

In Figure \ref{fig:nanogpt_lookahead_convergence}, we measure the validation loss at different lookahead positions during the learning rate decay phase when the learning rate is very small. Since the loss landscape changes rapidly near the local minima, our EMA lookahead direction ${\bf m}^t$ with large $\gamma$ retained the lookahead momentum from past iterations which is not applicable to the sharp local minima. This motivates the decaying schedule of $\beta_t$ and the early rest stage described in Section \ref{sec:beta_schedule}.

\subsection{Learning Rate Scaling with Batch Size}
\begin{figure}[htb!]
    \centering
    \includegraphics[width=0.35\linewidth]{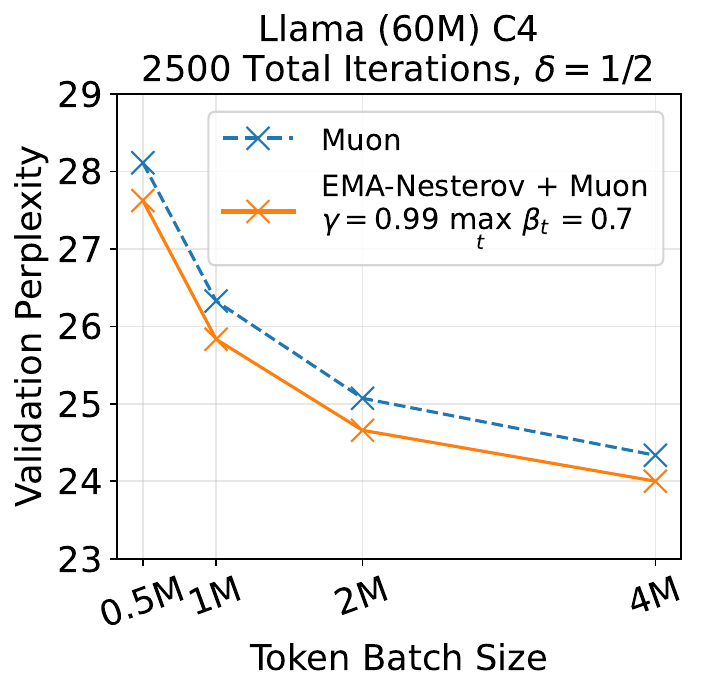}
    \includegraphics[width=0.35\linewidth]{figures/ema_nesterov_llama_60M_bs_0_25_scaling.pdf}
    \caption{Data scaling to total training budgets of $\ge \{20\times, 40\times, 80\times, 160\times\}$ tokens-per-parameter with increased token batch size $\{0.5{\rm M}, 1{\rm M}, 2{\rm M}, 4{\rm M}\}$ respectively and fixed 2500 total iterations. The learning rates are adapted under the power scaling rule: for tuned learning rate $\alpha$ on batch size $B$, the adapted learning rate $\alpha'$ for larger batch size $B' > B$ is $\alpha' = \alpha \cdot (B'/B)^{\delta}$ with the learning rate scaling exponent (Left) $\delta = 1/2$ and (Right) $\delta=1/4$.}
    \label{fig:lr_scaling_comp}
\end{figure}

In Figure \ref{fig:lr_scaling_comp}, we compare the two learning rate scaling rules under different scaling exponents. Compared to the usual scaling exponent $\delta = 1/2$ used in SGD \citep{hoffer2017train}, we tried alternative scaling exponent $\delta = 1/4$ and observed a better performance for Muon especially on larger batch sizes.

\subsection{Hyperparameter Sensitivity on Model Scaling} \label{app:sensitivity}

\begin{figure}[hbt!] 
    \centering
    \includegraphics[width=0.9\linewidth]{figures/llama_legend.pdf}\\
    \includegraphics[width=0.32\linewidth]{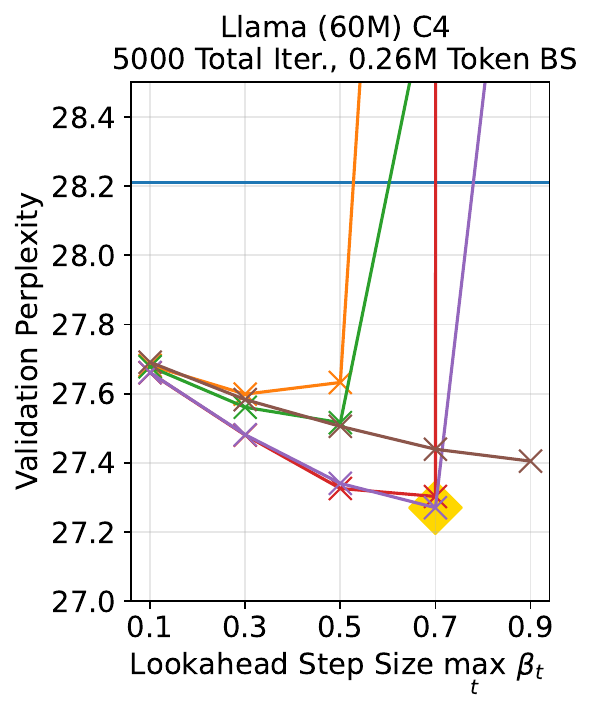}
    \includegraphics[width=0.32\linewidth]{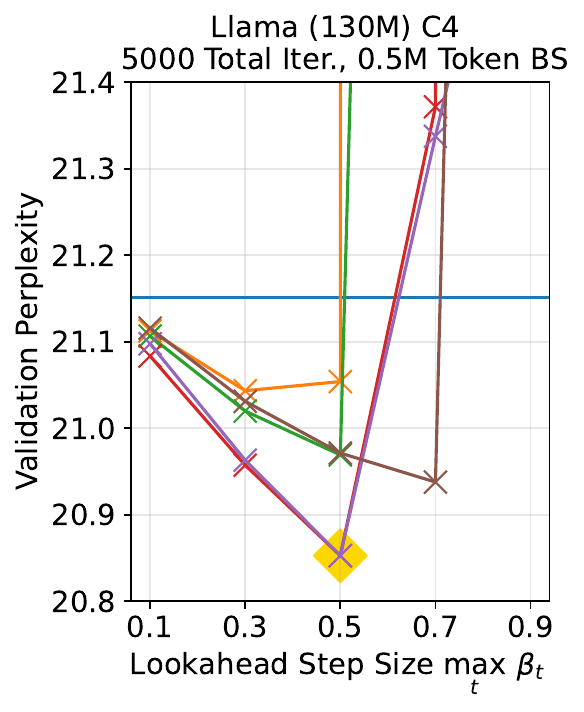}
    \caption{Sensitivity of hyperparameters for {\algname} + Muon across different model sizes on the Llama setup. The value of $\max_t \beta_t$ is searched in $\{0.1, 0.3, 0.5, 0.7, 0.9\}$ across the horizontal axis and the value of $\gamma$ is searched in $\{0.9, 0.95, 0.99, 0.995, 0.999\}$ across different colors. The learning rates are adapted under the power scaling rule: for tuned learning rate $\alpha$ on batch size $B$, the adapted learning rate $\alpha'$ for larger batch size $B' > B$ is $\alpha' = \alpha \cdot (B'/B)^{\delta}$ with $\delta = 1/2$.}
    \label{fig:llama_model_scaling_app}
\end{figure}

In Figure \ref{fig:llama_model_scaling_app} we provide additional results on the hyperparameter sweep for {\algname} when training the Llama model for 5000 iterations. Compared to the 2500 iterations setup in Figure \ref{fig:llama_model_scaling} where the optimal EMA rate $\gamma = 0.99$, the optimal EMA rate in Figure \ref{fig:llama_model_scaling_app} shifts to $\gamma = 0.995$. This indicates that the choice of $\gamma$ is determined by the temporal behavior of the training trajectory, such that the different training trajectories due to different total iterations and learning rate schedule lead to different optimal choice of $\gamma$.
\newpage
\subsection{Experiment Setting} \label{app:exp}
In experiment (I) training NanoGPT (124M) model on FineWeb dataset, our setting strictly follows an early version (\emph{track\_1\_short/2024-10-10\_Muon}) of the NanoGPT Track 1 benchmark, including the implementation of the Muon algorithm. A total of 6200 iterations is trained with batch size 512 on sequence length of 1024 tokens, approximately 0.5M token batch size. The learning rate is kept constant from the beginning and a linear decay to 0 for the last 1800 iterations is adopted. In Table \ref{tab:nanogpt_hp}, we summarize the hyperparameters of the base optimizers used in this setting. For {\algname}, we adopted the same $\beta_t$ scheduling of $T_w = 1800$ and $T_r = 6200 - 600 = 5600$ for all NanoGPT (124M) experiments. Since no learning rate warm-up is applied, we expect the training takes longer to enter the stable stage and therefore the slightly larger $T_w$. Meanwhile, as the learning rate decay phase is long, a late $T_r$ is adopted to accelerate the mid-to-late stage training. 
The time for each experiment run varies for different base optimizers, ranging from 24 minutes for Muon and 37 minutes for SOAP in Figure \ref{fig:nanogpt_ema_base_opts}.

We also conducted experiments using the NanoGPT Track 3 setup\footnote{NanoGPT Track 3 benchmark from \url{https://github.com/KellerJordan/modded-nanogpt/tree/master/records/track_3_optimization}.}, which differs from Track 1 by increasing the number of trainable parameters to 162M through detaching the parameters of input and output layer. For the base optimizers in Track 3 of Figure \ref{fig:nanogpt_ema_base_opts},
we adopted Muon from leaderboard entry \#12
and Aurora from pull request entry \#300. For the $\beta_t$ scheduling in {\algname}, we used $T_w = 500$, $T_r = 3150 - 750 = 2400$, $\max_t \beta_t = 0.6$ for Muon and $T_w = 300$, $T_r = 2900 - 950 = 1950$, $\max_t \beta_t =0.3$ for Aurora. We used $\gamma = 0.99$ for {\algname} in both algorithms.
\begin{table}[hbt!] 
    \centering
    \scalebox{0.82}{
    \begin{tabular}{l|c c c c c}
    \toprule 
     & Muon LR  & Adam LR & Muon Momentum & Adam 1st Momentum & Adam 2nd Momentum \\[0.05cm]
    \midrule
    Muon & $3.6 \times 10^{-4}$ & $3.6 \times 10^{-3}$ & 0.95 & 0.9 & 0.95\\
    NorMuon & $3.6 \times 10^{-4}$ & $3.6 \times 10^{-3}$ & 0.95 & 0.9 & 0.95\\
    \midrule
     & LR & 1st Momentum & 2nd Momentum & Precondition Frequency & - \\[0.05cm]
    \midrule
    SOAP & $10^{-3}$ & 0.9 & 0.95 & 10 & - \\
    Adam & $10^{-3}$ & 0.9 & 0.95 & - & - \\
    \bottomrule
    \end{tabular}
    }
    \caption{Hyperparameter values of the base optimizers used in NanoGPT Track 1 (Figure \ref{fig:nanogpt_ema_base_opts} and \ref{fig:nanogpt_ema_pessimistic_gpa_snoo}).}
    \label{tab:nanogpt_hp}
\end{table}

In experiment (II) training Llama models on C4 dataset, we used a sequence length of 256 tokens. Total training iterations and batch sizes vary between different experiment runs, as described in the title of Figure \ref{fig:llama_data_scaling}, \ref{fig:llama_model_scaling}, \ref{fig:lr_scaling_comp} and \ref{fig:llama_model_scaling_app}. 
The learning rate follows a warmup-stable-decay scheduling \citep{hu2024minicpm}, with the first 10\% of iterations performing linear warm-up, followed with staying constant in the middle 80\% of iterations and ends with an exponential decay to a minimum of 0.1$\times$ the maximum learning rate for the last 10\% of iterations. 
For the scheduling of $\beta_t$ in {\algname}, when the total iteration is 2500, we used $T_w = 750$ and $T_r = 2500 - 500 = 2000$. When the total iteration is 5000, it is relatively scaled up to $T_w = 1500$ and $T_r = 5000 - 1000 = 4000$. This $\beta_t$ schedule roughly follows the learning rate schedule with an additional safe margin, i.e., $\beta_t$ used a 30\% of early iterations to warm-up (i.e., $\beta_t=0$ for $t \leq 0.3\cdot  T$) and uses a 20\% of late iterations to rest (i.e., $\beta_t = 0$ for $t > 0.8\cdot T$). 
We only used Muon\footnote{For all Llama experiments, we adopted the open-sourced implementation of Muon in \url{https://github.com/KellerJordan/Muon}.} as the base optimizer in this setting and it used a momentum of 0.95 for intermediate layers, a 1st and 2nd order momentum of 0.9, 0.95 for first/last layers, see Algorithm \ref{alg:muon_adam} for a detailed description. We summarize the remaining hyperpameters for each figure in Table \ref{tab:llama_hp}. The time for experiment varies across model sizes, ranging from 40 mintues for 60M model in Figure \ref{fig:llama_model_scaling} to 24 hours for 1B model in Table \ref{tab:all_llamas}.
\newpage
\begin{table}[hbt!] 
    \centering
    \scalebox{0.9}{
    \begin{tabular}{l|c c c c c}
    \toprule 
     & Model Size & Total Iterations  & Token Batch Size & Muon LR & Adam LR \\[0.05cm]
    \midrule
    Figure \ref{fig:llama_model_scaling} & 60M & 2500 & 0.5M & $3\times 10^{-2}$ & $3\times 10^{-2}$ \\
    Figure \ref{fig:llama_model_scaling} & 130M & 2500 & 1M & $4.24\times 10^{-2}$ & $4.24\times 10^{-2}$ \\
    Figure \ref{fig:llama_model_scaling} & 350M & 2500 & 3M & $7.35\times 10^{-2}$ & $7.35\times 10^{-2}$ \\
    \midrule
    Figure \ref{fig:lr_scaling_comp} ($\delta=1/2$) & \multirow{4}{*}{60M} & \multirow{4}{*}{2500} & 0.5M & $3\times 10^{-2}$ & $3\times 10^{-2}$ \\
    Figure \ref{fig:lr_scaling_comp} ($\delta=1/2$) &  &  & 1M & $4.24\times 10^{-2}$ & $4.24\times 10^{-2}$ \\
    Figure \ref{fig:lr_scaling_comp} ($\delta=1/2$) &  &  & 2M & $6\times 10^{-2}$ & $6\times 10^{-2}$ \\
    Figure \ref{fig:lr_scaling_comp} ($\delta=1/2$) &  &  & 4M & $8.49\times 10^{-2}$ & $8.49\times 10^{-2}$ \\
    \midrule
    Figure \ref{fig:llama_data_scaling}, \ref{fig:lr_scaling_comp} ($\delta=1/4$) & \multirow{4}{*}{60M} & \multirow{4}{*}{2500} & 0.5M & $3\times 10^{-2}$ & $3\times 10^{-2}$ \\
    Figure \ref{fig:llama_data_scaling}, \ref{fig:lr_scaling_comp} ($\delta=1/4$) &  & & 1M & $3.56\times 10^{-2}$ & $3.56\times 10^{-2}$ \\
    Figure \ref{fig:llama_data_scaling}, \ref{fig:lr_scaling_comp} ($\delta=1/4$) &  & & 2M & $4.24\times 10^{-2}$ & $4.24\times 10^{-2}$ \\
    Figure \ref{fig:llama_data_scaling}, \ref{fig:lr_scaling_comp} ($\delta=1/4$) &  & & 4M & $5.05\times 10^{-2}$ & $5.05\times 10^{-2}$ \\
    \midrule
    Figure \ref{fig:llama_data_scaling} ($\delta=1/4$) & \multirow{4}{*}{130M} & \multirow{4}{*}{2500} & 1M & $4.24\times 10^{-2}$ & $4.24\times 10^{-2}$ \\
    Figure \ref{fig:llama_data_scaling} ($\delta=1/4$) &  & & 2M & $5.05\times 10^{-2}$ & $5.05\times 10^{-2}$ \\
    Figure \ref{fig:llama_data_scaling} ($\delta=1/4$) &  & & 4M & $6\times 10^{-2}$ & $6\times 10^{-2}$ \\
    Figure \ref{fig:llama_data_scaling} ($\delta=1/4$) &  & & 8M & $7.14\times 10^{-2}$ & $7.14\times 10^{-2}$ \\
    \midrule
    Figure \ref{fig:llama_data_scaling} ($\delta=1/4$) & \multirow{4}{*}{350M} & \multirow{4}{*}{2500} & 3M & $3\times 10^{-2}$ & $3\times 10^{-2}$ \\
    Figure \ref{fig:llama_data_scaling} ($\delta=1/4$) &  & & 6M & $3.57\times 10^{-2}$ & $3.57\times 10^{-2}$ \\
    Figure \ref{fig:llama_data_scaling} ($\delta=1/4$) &  & & 12M & $4.24\times 10^{-2}$ & $4.24\times 10^{-2}$ \\
    Figure \ref{fig:llama_data_scaling} ($\delta=1/4$) &  & & 25M & $5.05\times 10^{-2}$ & $5.05\times 10^{-2}$ \\
    \midrule
    Figure \ref{fig:llama_model_scaling_app} & 60M & \multirow{2}{*}{5000} & 0.26M & $2.12\times 10^{-2}$ & $2.12\times 10^{-2}$ \\
    Figure \ref{fig:llama_model_scaling_app} & 130M &  & 0.5M & $3\times 10^{-2}$ & $3\times 10^{-2}$ \\
    \midrule
    Table \ref{tab:all_llamas} & 1B & 5000 & 4M & $5\times 10^{-2}$ & $5\times 10^{-2}$ \\
    \bottomrule
    \end{tabular}
    }
    \caption{Hyperparameter values of the base optimizers used in Figure \ref{fig:llama_model_scaling}, \ref{fig:llama_data_scaling}, \ref{fig:lr_scaling_comp}, \ref{fig:llama_model_scaling_app} and Table \ref{tab:all_llamas}. The learning rates are adapted under the power scaling rule: for tuned learning rate $\alpha$ on batch size $B$, the adapted learning rate $\alpha'$ for larger batch size $B' > B$ is $\alpha' = \alpha \cdot (B'/B)^{\delta}$ with a default choice of $\delta = 1/2$.}
    \label{tab:llama_hp}
\end{table}
\newpage
\subsection{Muon Optimizer}\label{app:muon}
For completeness, we provide the exact open-sourced implementation of Muon\footnote{See the open-sourced implementation here \url{https://github.com/KellerJordan/Muon}.} in Algorithm \ref{alg:muon_adam}. Unless otherwise specified, we do not apply weight decay in the base optimizer, i.e., we use $\lambda = 0$ as default. We denote $\odot$ as the element-wise product operator.

\begin{algorithm}
\caption{Muon}
\label{alg:muon_adam}
\begin{algorithmic}[1]
\STATE \textbf{Input:} Muon learning rates $\alpha_t > 0$, Adam learning rates $\widehat{\alpha}_t > 0$, weight decay $\lambda \ge 0$, Muon momentum $\phi \in [0, 1)$, Adam momentum $\phi_1, \phi_2\in [0, 1)$.
\STATE \textbf{Initialize:} Model parameters $\bx^0 = [\bx_1^0, ..., \bx_L^0]$ for $L$ layers s.t. $\bx^0_{\ell} \in \mathbb{R}^{M_\ell \times N_\ell}$ with input dimension $M_\ell$ and output dimension $N_\ell$. Muon gradient momentum ${\bf v}^0 = [{\bf v}_2^0, ..., {\bf v}_{L-1}^0] = {\bf 0}$. Adam 1st and 2nd order gradient momentum ${\bf w}^0 = [{\bf w}_1^0, ..., {\bf w}_L^0] = {\bf 0}$, $\widehat{{\bf w}}^0 = [\widehat{\bf w}_1^0, ...,\widehat{\bf w}_L^0] = {\bf 0}$.

\FOR{$t = 0, ..., T-1$}
\FOR{$\ell = 1, ..., L$}
    \IF{$\ell \in [2, L-1]$ and $M_\ell > 1$ and $N_\ell > 1$}
        \STATE $\alpha_{t, \ell} = \alpha_t \cdot  \sqrt{\max(1, N_\ell/M_\ell)}$ 
        \STATE \label{alg_step:muon_double_ema_1} $\bx^{t+1}_{\ell} = (1 - \alpha_t \lambda) \bx^t_{\ell} - \alpha_{t, \ell} \cdot \text{Newton-Schulz5}(\phi^2 {\bf v}^t_{\ell} + (1-\phi^2) \nabla f(\bx^t; \xi^t)_{\ell} )$
        \STATE  \label{alg_step:muon_double_ema_2} ${\bf v}^{t+1}_{\ell} = \phi {\bf v}^t_{\ell} + (1 - \phi) \nabla f(\bx^t; \xi^t)_{\ell}$
    \ELSE 
        \STATE ${\bf w}^{t+1}_{\ell} = \phi_1 {\bf w}^{t}_{\ell} + (1 - \phi_1) \nabla f(\bx^t; \xi^t)_\ell$
        \STATE $\widehat{\bf w}^{t+1}_{\ell} = \phi_2 \widehat{\bf w}^{t}_{\ell} + (1 - \phi_2) \nabla f(\bx^t; \xi^t)_\ell \odot f(\bx^t; \xi^t)_\ell$
        \STATE $\bx^{t+1}_{\ell, ij} = (1 - \widehat{\alpha}_{t} \lambda)\bx^t_{\ell, ij} - \widehat{\alpha}_{t} \cdot \frac{{\bf w}^{t+1}_{\ell, ij} / (1 - \phi_1^t)}{\sqrt{\widehat{\bf w}^{t+1}_{\ell, ij} / (1 - \phi_2^t)} + 10^{-10} }$\quad for \quad $i=1,..., M_\ell$, $j=1,..., N_\ell$. 
    \ENDIF
\ENDFOR
\ENDFOR
\STATE \textbf{Output:} last iterate solution $\bx^{T}$.
\end{algorithmic}
\end{algorithm}

Notice that Algorithm \ref{alg:muon_adam} relies on the following 5-steps Newton-Schulz iteration algorithm to approximately normalize the spectral values of the update direction.
\begin{algorithm}
\caption{Newton-Schulz5}
\begin{algorithmic}[1]
\STATE \textbf{Input:} ${\bf X} \in \mathbb{R}^{M\times N}$.
\IF{$N > M$}
    \STATE $\widehat{\bf X} = {\bf X}^\top$
\ELSE
    \STATE $\widehat{\bf X} = {\bf X}$
\ENDIF

\STATE ${\bf X}^0 = \widehat{\bf X} / (\| \widehat{\bf X} \|_F + 10^{-7})$
        \FOR{$t = 0$ \textbf{to} $4$} 
            \STATE ${\bf A}^t = {\bf X}^t({\bf X}^t)^\top $
            \STATE ${\bf B}^t = ( 2.0315 \cdot {\bf A}^t-4.7750 \cdot {\bf I} ){\bf A}^t$
            \STATE ${\bf X}^{t+1} = (3.4445 \cdot {\bf I} + {\bf B}^t) {\bf X}^t$
        \ENDFOR
\IF{$N > M$}
    \STATE $\widetilde{\bf X} = ({\bf X}^5)^\top$
\ELSE
    \STATE $\widetilde{\bf X} = {\bf X}^5$
\ENDIF
\STATE \textbf{Output:} $\widetilde{\bf X}$.
\end{algorithmic}
\end{algorithm}

We remark that in the source code of Muon, the convex combination between ${\bf v}_\ell^t$ and $\nabla f(\bx^t; \xi^t)_\ell$ in line \ref{alg_step:muon_double_ema_1} of Algorithm \ref{alg:muon_adam} is referred to as using "Nesterov". In contrary to the description of its source code, we demonstrate explicitly that line \ref{alg_step:muon_double_ema_1} of Algorithm \ref{alg:muon_adam} is still based on EMA of stochastic gradient with different EMA rate. In particular, ${\bf v}_\ell^t$ is an EMA of rate $\phi$ for $\{\nabla f(\bx^r; \xi^r)\}_{r=0}^{t-1}$ which is combined on-the-fly with the current stochastic gradient $\nabla f(\bx^t; \xi^t)_\ell$ of a larger weighting $1 - \phi^2$ because $1 - \phi^2 > 1 - \phi$ for $\phi \in (0,1)$. The larger weighting $1 - \phi^2$ controls a bias-variance tradeoff as the current stochastic gradient is unbiased to the current solution $\bx^t$.

\end{document}